%% file: ms.tex
\pdfoutput=1
\documentclass[10pt,twocolumn,letterpaper]{article}

\usepackage[pagenumbers]{cvpr}

\usepackage{makecell}
\usepackage{booktabs, makecell, graphicx}
 \usepackage{multirow}

\input{preamble_arxiv}
\definecolor{cvprblue}{rgb}{0.21,0.49,0.74}
\usepackage[pagebackref,breaklinks,colorlinks,allcolors=cvprblue]{hyperref}

\def\confName{CVPR}
\def\confYear{2027}

\title{\OURS{}: Agentic Evolutionary Search for Principled Rearrangement of 3D Indoor Scenes}

\author{Chandan Yeshwanth \qquad Angela Dai\\[2mm]
Technical University of Munich
}

\input{constants_arxiv}

\input{commands_arxiv}
\begin{document}
\twocolumn[{%
\renewcommand\twocolumn[1][]{#1}%
\maketitle
\begin{center}
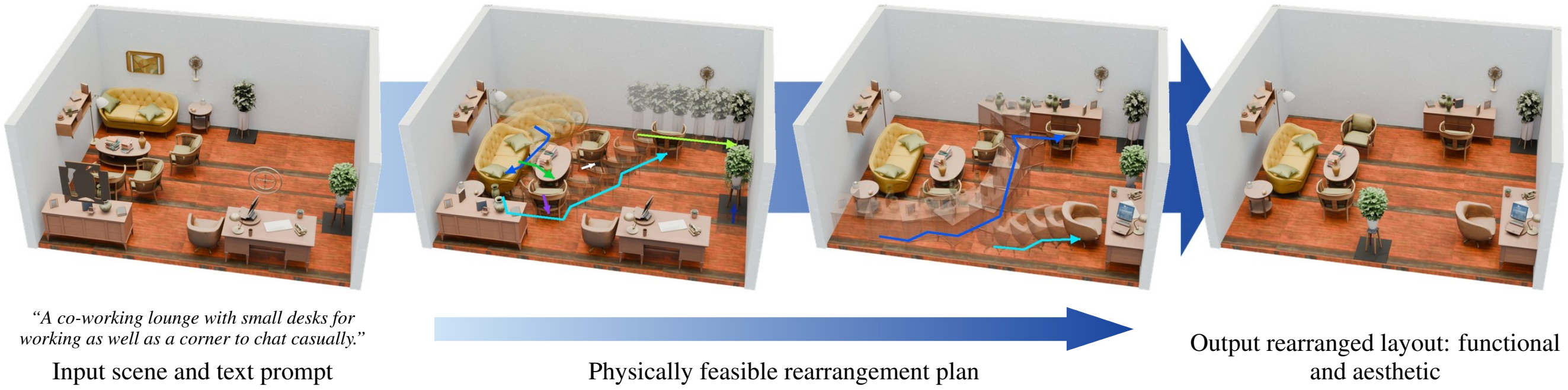

    \centering
    \captionsetup{type=figure}
    \input{sec_arxiv/fig_teaser}
    \captionof{figure}{\OURS{} rearranges indoor scenes through an agentic evolutionary search that systematically explores the solution space of interior design, giving physically feasible, functional and aesthetic rearrangements that adhere to the input prompt and to common interior design principles.}
    \label{fig:teaser}
\end{center}%
}]
\input{sec_arxiv/0_abstract}
\input{sec_arxiv/1_intro}
\input{sec_arxiv/2_relatedwork}
\input{sec_arxiv/method}
\input{sec_arxiv/experiments}
\input{sec_arxiv/conclusion}
{
    \small
    \bibliographystyle{ieeenat_fullname}
    \bibliography{main_arxiv}
}
\input{sec_arxiv/suppl}

\end{document}

%% file: preamble_arxiv.tex
\makeatletter
\let\oldparagraph\paragraph
\renewcommand{\paragraph}{%
  \if@nobreak\else\vspace{-0.5cm}\fi
  \oldparagraph}
\makeatother

%% file: constants_arxiv.tex
\newcommand{\OURS}{SceneDesigner3D}

\newcommand{\ImprovementPA}{5.1}
\newcommand{\ImprovementDP}{2.1}
\newcommand{\ImprovementComputed}{15.2}

\newcommand{\NumScenesSAGE}{33}
\newcommand{\NumScenesAblations}{15}
\newcommand{\NumScenesSAGEOpus}{31}
\newcommand{\NumScenesSceneSmith}{25}

\newcommand{\NumBaselines}{6}

\newcommand{\UserStudyNumUsers}{15}
\newcommand{\UserStudyNumPairs}{18}
\newcommand{\UserStudyNumSamples}{21}

%% file: sec_arxiv/fig_teaser.tex
{\setlength{\tabcolsep}{0pt}%
\begin{tabular}{@{}ccccccc@{}}
\multicolumn{7}{@{}l@{}}{\includegraphics[width=\textwidth]{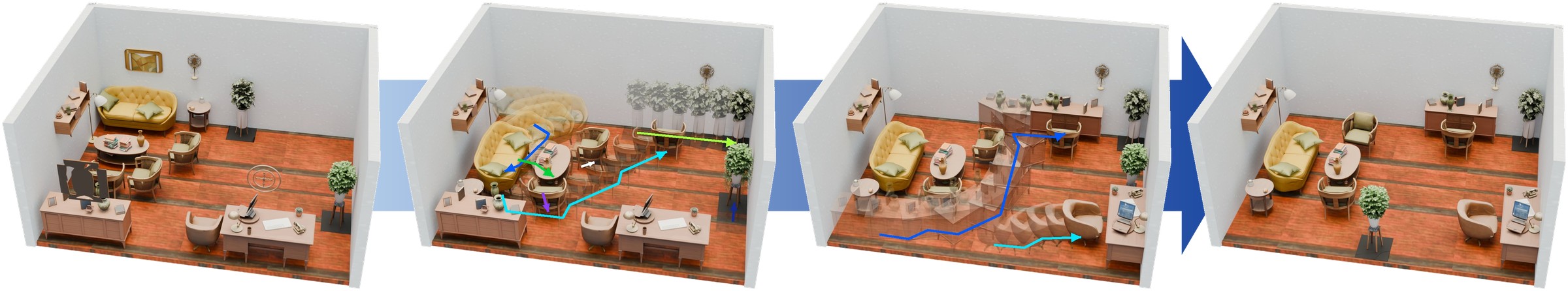}} \\[1pt]
\parbox[b]{0.2440\textwidth}{\centering\fontsize{6pt}{6.6pt}\selectfont \textit{``A co-working lounge with small desks for working as well as a corner to chat casually.''}\\[2pt]\fontsize{8pt}{9pt}\selectfont Input scene and text prompt} & \hspace*{0.0080\textwidth} & \multicolumn{3}{c}{\parbox[b]{0.4960\textwidth}{\centering\includegraphics[width=0.4464\textwidth]{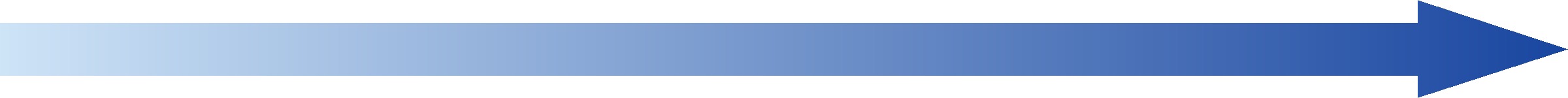}\\[0pt]\fontsize{8pt}{9pt}\selectfont Physically feasible rearrangement plan}} & \hspace*{0.0080\textwidth} & \parbox[b]{0.2440\textwidth}{\centering\fontsize{8pt}{9pt}\selectfont Output rearranged layout: functional and aesthetic} \\
\end{tabular}}%

%% file: sec_arxiv/0_abstract.tex
\begin{abstract}
Scene rearrangement is an essential task for 3D content creation applications, requiring generation of diverse, aesthetic and functional scene rearrangements. We present \OURS{}, an agentic evolutionary search method for 3D indoor scene rearrangement that systematically explores the design space of 3D scenes and generates multiple solutions according to common interior design principles. Given an input scene and a text prompt, \OURS{} represents rearrangements as editable programs and evolves a population of candidate layouts. Starting from distinct design ideas, a VLM agent mutates the programs by refining promising candidates, exploring new rearrangement strategies or redesigning from scratch, while verifying layouts for physical validity. Proposals are scored by a VLM judge for prompt adherence and satisfaction of design principles such as balance, zoning and emphasis, as well as computed principles such as circulation and clutter. An archive of top-performing, structurally varied candidates further increases diversity. Experiments on the SAGE-10k and SceneSmith datasets show that \OURS{} improves prompt adherence and computed design principle scores over baselines by up to \ImprovementPA{}\% and \ImprovementComputed{}\%, respectively. Code will be made publicly available.
\enlargethispage{\baselineskip}
\end{abstract}

%% file: sec_arxiv/1_intro.tex
\section{Introduction}
\label{sec:intro}
3D indoor scene rearrangement is a central task for content creation applications such as interior design, games and movies, as well as embodied AI systems that need to plan and interact with the world around them. Given an input 3D indoor scene, the task is to rearrange the objects in the scene following the input prompt while keeping the scene physically plausible and functional without collisions or unusable objects. This task is challenging for several reasons: the solution space is combinatorial over tens to hundreds of input objects, and moving one object can affect the functionality and aesthetics of the scene both locally and globally. Moreover, the prompt, functionality and aesthetics often compete with each other and must be traded off. There are typically many desirable solutions, and committing to a single solution can miss better ones. Generating a diverse set of high-quality solutions can enable users and interior designers to choose the one that best fits their needs.

Previous work has shown notable results on the rearrangement task by leveraging the strong semantic priors present in LLMs (large language models) and VLMs (vision-language models) to generate hierarchical scene edits \cite{el2025scanedit}, generate code with optimizable constraints \cite{sun2025layoutvlm} or other structured formats \cite{feng2023layoutgpt}. They produce plausible rearrangements but optimize, at most, local constraints (``next to", ``facing"). However, functionally usable and aesthetic scenes require global principles such as clear walking paths, reachability, division of space into zones and visual balance among objects \cite{ching2012interior, merrell2011interactive}. These principles depend on the overall layout and cannot be captured by local constraints. Further, existing methods are designed to generate a single rearranged layout following the input text prompt, and do not systematically search the space of possible solutions to optimize for such principles.

In contrast, we propose an evolutionary approach to scene rearrangement in order to achieve both adherence to the input text prompt as well as satisfaction of common interior design principles. Evolving a diverse set of rearrangement proposals based on different high-level functional and aesthetic design seeds allows us to search the interior design space systematically and move towards more functional and aesthetically pleasing scene arrangements. Evolutionary methods have shown considerable progress in solving mathematical and programming tasks \cite{novikov2025alphaevolve, lange2026shinkaevolve, qu2026coral}, but their application to scene-level 3D generation and editing tasks remains limited \cite{kan2017automated, ran2026scenemosaic}.

To generate such functional and aesthetic scenes, we propose \OURS{}, an evolutionary search method built on top of frontier VLMs that iteratively mutates and evaluates rearrangements of an input scene. The rearrangement is represented as a mutable program that takes the original scene as input and moves its objects. The program is updated by a VLM mutator agent that can propose and validate programs before sending them to a VLM judge for evaluation. The mutator only edits programs, and the evolutionary search maintains a population of valid layouts and selects promising candidates to mutate further. We decompose the input prompt into a few verifiable
requirements, and use ten abstract design principles from the interior
design literature, such as zoning, balance and emphasis \cite{merrell2011interactive, ching2012interior}. Each proposal is rendered from multiple viewpoints, and a VLM judge compares it to its parent on every requirement and principle. It returns relative preferences, absolute scores, and concrete reasoning grounded in the layout as feedback to the mutator. Principles that can be measured geometrically, such as circulation, focal point and clutter, are computed numerically from the layout. The computed scores give the mutator precise feedback across iterations, and together with the judge scores they determine the fitness.

We design several mechanisms to grow a diverse set of candidate rearrangements, such as proposing various initial design seeds for the layouts and refining them across islands, and an archive \cite{mouret2015illuminating} of high-quality layouts that serve as inspiration for new proposals. Additionally, we enable the VLM mutator to choose from different mutation strategies: exploring initial design seeds, localized edits to refine existing proposals, and generation of large mutations to explore rearrangements dissimilar to the parent proposal.

We evaluate \OURS{} and several state-of-the-art scene editing and generation baselines on two agentically generated synthetic indoor scene datasets: SAGE-10k \cite{xia2026sage} and SceneSmith \cite{pfaff2026scenesmith}, as well as an improved version of SAGE-10k generated using Claude Opus 5.5. Our method improves upon the baselines in all settings in terms of prompt adherence, design principle scores and diversity.

\noindent
To summarize, our contributions are:
\begin{itemize}
    \item an agentic evolutionary search method for 3D scene rearrangement that represents the rearrangement as an editable program, and mutates the programs with a VLM agent that drafts and validates candidates
    \item an evaluator that verifies commonly used design principles, consisting of a VLM judge that assesses abstract design principles such as balance and emphasis, along with geometric computation of measurable principles such as circulation and clutter
    \item diversity mechanisms that generate structurally different rearrangements, including multiple high-level seed ideas, a mutator that can switch between seed exploration, large rearrangements and fine-grained refinements, and an archive of high-quality layouts capturing varied anchor-object placements

\end{itemize}

%% file: sec_arxiv/2_relatedwork.tex
\section{Related Work}
\label{sec:related_work}

\subsection{LLM-based 3D Scene Generation and Editing}
Previous methods such as ScanEdit \cite{el2025scanedit} and HOG-Layout \cite{jiang2026hog} use LLMs for planning primarily local scene edits hierarchically or in groups. Edit-As-Act \cite{noh2026edit} uses goal-regressive planning to plan local edits. LayoutVLM \cite{sun2025layoutvlm} uses a VLM for generating a programmatic representation of a scene along with optimization constraints. LayoutGPT \cite{feng2023layoutgpt} and HDSL \cite{li2026hdsl} use LLMs to generate or edit the scene in structured formats. EditRoom \cite{zheng2025editroom} performs a limited set of edits through scene graph diffusion. SceneAssistant \cite{luo2026sceneassistant} uses an agentic refinement loop for physically validated scene generation.  These methods do not optimize for overall functionality and aesthetics of the generated layouts. In contrast, we use an agentic evolutionary search that proposes diverse layout candidates using an LLM and VLM while explicitly optimizing for diversity and design principles.

\subsection{Agentic Evolutionary Methods}
ShinkaEvolve \cite{lange2026shinkaevolve} adapts standard evolutionary methods for sample efficiency using LLM agents, based on AlphaEvolve which generates thousands of proposals \cite{novikov2025alphaevolve}. Similarly, ThetaEvolve \cite{wang2025thetaevolve} and FrugalEvo \cite{chen2026frugalevo} reduce the cost of the evolution search through cheaper LLMs or make use of larger program databases. These methods have primarily been used for programming and algorithmic tasks, and their application to 3D indoor scenes is limited. We base our method on ShinkaEvolve, and add diversity mechanisms, agentic mutation and VLM evaluation of design principles for the task of 3D scene rearrangement.

\subsection{Evolution and Principled 3D Scene Generation}
Previous methods have optimized layouts for design principles with evolutionary algorithms \cite{kan2017automated, akase2013automatic} or greedy cost minimization
\cite{kan2018automatic}. SceneMosaic \cite{ran2026scenemosaic} generates a set of scene rearrangements with limited diversity by performing an agentic refinement loop on the input scene, while we perform a full evolutionary search that systematically explores the interior design space.
LayoutEnhancer \cite{leimer2022layoutenhancer} uses an ergonomic loss that optimizes for limited design principles on top of a generative model, but does not search through the design space.
Make it Home \cite{yu2011make} infers object relationships from example scenes, and optimizes for a fixed set of design principles such as accessibility and visibility. Unlike these methods, \OURS{} conditions on text prompts, is readily extensible to new design principles and does not need hand-tuned differentiable cost functions.

\subsection{Scene Rearrangement Planning}
Rearrangement planning methods such as
SceneMover \cite{wang2020scene}, ESRP \cite{chen2026embodied}, PEARL \cite{wang2021pearl}, PLATO \cite{wang2024mastering} and HAISOR \cite{sun2024haisor} take the start and end states of the scene as input, and optimize with reinforcement learning for shortest paths. In contrast, we show an extension of \OURS{} that jointly plans moves during evolution.

%% file: sec_arxiv/method.tex
\section{Method}
\label{sec:method}

\begin{figure*}[ht!]
\centering
\includegraphics[width=\textwidth]{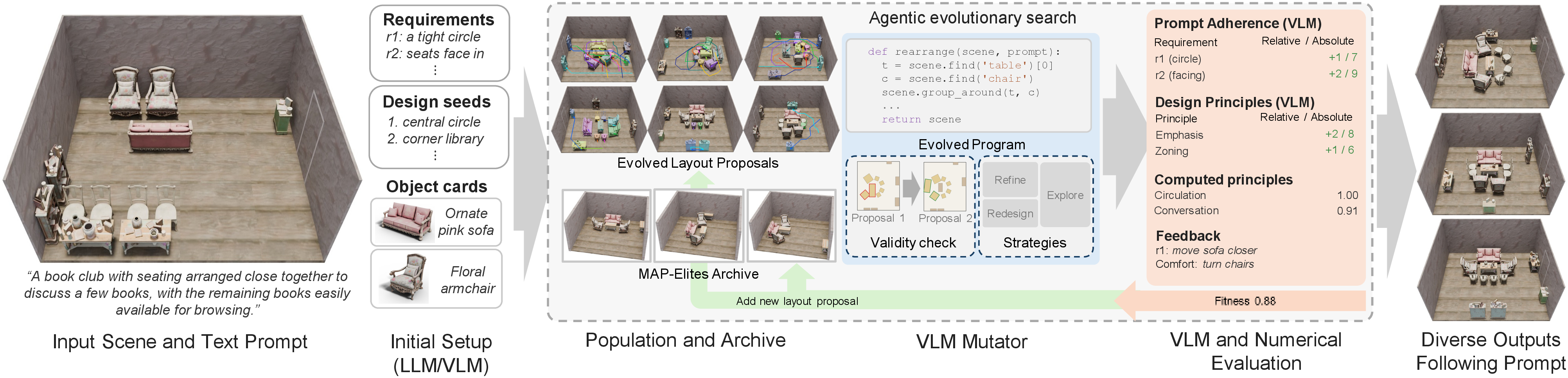}
\caption{\textbf{Method overview.} Given an input 3D scene and a text prompt, \OURS{} evolves a population of programs that rearrange its objects. The prompt is decomposed into requirements, design seeds are initialized and objects are described by a VLM. An agentic evolutionary search explores the design space through various strategies (colored paths: object movement relative to parent). Each candidate is judged by a VLM against its parent on the prompt requirements and design principles, and combined with numerically computed principles into its fitness. An archive keeps diverse candidates as inspiration for future iterations, giving diverse layouts that follow the prompt and design principles.}
\label{fig:method_overview}
\end{figure*}

\paragraph{Problem Formulation}
Given an input text prompt $T$ and a 3D scene $S=(L,\{o_i\}_{i=1}^{N})$ with objects $o_i$, the task of scene rearrangement is to edit the positions and orientations of the objects according to $T$. $L$ is the layout of the scene, represented as a set of polygons of the floor, walls, doors and windows. Each object $o_i=(c_i, m_i, \mathbf{p}_i, \theta_i)$ is defined by its semantic class $c_i$, 3D mesh $m_i$ in canonical orientation, 3D position $\mathbf{p}_i$ and rotation about the vertical axis $\theta_i$. The output is a scene $S'$ containing the same set of objects with only their positions and orientations changed, such that $S'$ is physically valid without collisions and floating objects, adheres to $T$ and satisfies interior design principles such as zoning, balance, circulation and low clutter.

\paragraph{Overview of \OURS{}}
We formulate rearrangement as an evolutionary search over programs that edit the scene (\cref{fig:method_overview}). \OURS{} evolves a population of such programs by repeating parent selection, mutation and evaluation steps. In each iteration, \OURS{} selects a parent to mutate and a mutation strategy from a fixed set of strategies, and produces a new candidate layout through an agentic session. The layout is then evaluated by a VLM for prompt adherence and abstract design principles, as well as numerically for geometric design principles. These three evaluations are combined into a single fitness score, and unique high-fitness layouts are stored in an archive. Finally, the highest-fitness layouts are selected.

\subsection{Scene and Program Representation}
\label{sec:program}
The rearrangement is represented as a single Python function \texttt{rearrange(scene, prompt)} that operates on the input scene $S$. The \texttt{rearrange} function queries and edits the scene through a fixed scene API,  which allows it to query the current scene geometrically for object sizes, their front directions and distances to walls, and edit the scene through functions such as \texttt{move} and \texttt{put\_on}. The API also contains other functions for finding free spots and grouping objects together. Only the \texttt{rearrange} function is evolved.
In addition to the input object arrangement, the program has access to \textit{object cards}: a fixed set of useful descriptors of each object including appearance, material and functionality, extracted by a VLM from multiview renders of the objects.

\paragraph{Prompt Requirements} The input prompt is decomposed into a set of fine-grained requirements $ \mathcal{R}=\{r_j\}$ by an LLM once before the evolution starts. This allows the evaluation step to focus on different parts of the prompt that need to be satisfied in isolation.

\subsection{Population and Parent Selection}
\label{sec:selection}
We initialize the population of candidate layouts using $K$ \textit{design seeds}, which are high-level ideas output by an LLM based on the input scene arrangement. The seeds place large objects in different regions of the scene. They only decide different initial directions for the evolution and can be freely updated in future iterations. Evolution is performed in $M$ different islands. In every future iteration, a parent $i$ is sampled for mutation using the \textit{weighted} strategy from ShinkaEvolve~\cite{lange2026shinkaevolve}, with a probability proportional to
\begin{equation}
w_i=\sigma\!(\lambda(\,a_i-\operatorname{med}(a){}))\,\frac{1}{1+n_i},
\end{equation}
where $a_i$ is the fitness of program $i$, $\operatorname{med}(a)$ is the median fitness, $\sigma$ is the sigmoid function, $n_i$ is the number of children of program $i$ and $\lambda$ controls the sharpness of the sigmoid. This weighted score controls the balance between exploitation of high-fitness proposals and exploration of parents with few children.

\paragraph{Mutation Strategies}
We further balance between exploration and exploitation during the mutation step by selecting from different high-level mutation strategies. First, we \textit{refine} the chosen parent if it has high fitness and previous refinements increased its fitness, keeping most of its coarse layout and making smaller local edits based on the weakest judged requirements and principles from previous iterations. Second, we choose to \textit{explore} new rearrangements by picking from a fixed sub-list of options:  \begin{enumerate*}[label=(\arabic*), itemjoin={{, }}, itemjoin*={{, and }}]
    \item re-seed the design by generating a new high-level idea
    \item make a large change by moving large objects
    \item move only objects that have not moved so far
    \item move 1--2 large objects to regions they have not occupied so far and rebuild around them
    \item attempt to exchange the positions of two large objects
    \item cross over the parent program with the inspiration program given as input context.
  \end{enumerate*}
The mutator agent selects one choice, given the number of attempts of each choice and how many times it led to a new best layout or a new archive cell.
The third option is a scheduled \textit{redesign} of the scene which must re-place a large fraction of the objects in the scene with a sufficient change in the scene-graph, constructed by inferring edges between nearby objects.

\subsection{Agentic Mutation}
\label{sec:mutation}
We send the selected parent layout to a VLM agent, which edits its program to generate a new child layout.
The agent operates in an isolated workspace, where it can create and run Python scripts, read provided input files and write to files. The agent is prompted with several inputs: a detailed description of the rearrangement task, the input scene and generated requirement list, the parent layout and text feedback from the previous evaluation, the current mutation strategy and mode, and an inspiration program sampled over all islands. The feedback describes the current fulfillment of each requirement and design principle and reasons for each rating.
The agent has access to several tools and context files, including the scene API, functions to compute the numerical design principles, renders of the parent layout, a top-down labelled plan indicating object locations in 2D, object cards and renders of objects. It can also access a \textit{test tool} that executes a candidate layout and reports colliding meshes, computed design principle scores and comparisons to existing layouts. The agent uses these tools and context files through a multi-turn session.

\subsection{Evaluation and Fitness}
\label{sec:fitness}
The evaluation step checks the validity and quality of the generated scene, finds unsatisfied prompt requirements and design principles using a VLM as well as numerically computed design principles, and packages these as concise feedback to following mutation steps. The prompt adherence and principles scores are combined together into a single fitness score, which is used to sample parents and maintain an archive of high-quality layouts.

\paragraph{Validity and uniqueness} The first check ensures that objects are within the scene bounds, do not collide or float or block doors and windows. The evaluation is stopped at this stage for invalid and near-duplicate proposals. Valid layouts are then rendered from several fixed viewpoints, which are judged by the VLM in the next steps.

\paragraph{Prompt Adherence}
The VLM judge uses renders of the child and parent layouts, the input text prompt $T$ and the requirements set $\mathcal R$, and scores every requirement $r\in\mathcal R$ with a preference on a $5$-point scale,  $q_r\in\{-2,\dots,2\}$ relative to the parent layout, predicting whether adherence has increased, decreased or remained the same. Additionally, it outputs an absolute adherence score for the whole prompt $s_T \in [0,10]$, as well as short reasoning. This allows fine-grained feedback both in isolation and in relation to the parent.
The preferences are averaged over the requirements, scaled by a fixed factor $\alpha$ and added to the rating of the parent $A$ to obtain the rating of the child $B$:
\[\rho_T(B)=\rho_T(A)+\frac{\alpha}{|\mathcal R|}\sum_{r\in\mathcal R} q_r\]
The ratings are initialized at $1$ for the input scene and updated for each child layout. In this way, small but consistent improvements can be measured.

\paragraph{Abstract Design Principles}
We incorporate ten abstract design principles that are commonly used in the literature \cite{ching2012interior, merrell2011interactive}, specified as short descriptions with examples: \begin{enumerate*}[label=(\arabic*), itemjoin={{, }}, itemjoin*={{, and }}]
\item \textit{zoning}: separate zones for different activities
\item \textit{floating and anchoring}: object groups may be anchored to a single object or free-standing
\item \textit{alignment} of objects to the architectural features of the scene
\item \textit{balance and proportion} of objects
\item \textit{emphasis} on a single focal point
\item \textit{composition}: intentional symmetry, asymmetry or a radial arrangement
\item \textit{comfort} through visibility of the scene and placement of objects
\item \textit{harmony and contrast} of different object styles
\item \textit{functionality} of every object being ensured
\item \textit{order} of objects such that they satisfy a meaningful purpose
\end{enumerate*}.
These descriptions are identical to the ones given to the mutation agent for a shared notion of quality. Similar to the prompt adherence evaluation, the VLM evaluates the design principles and outputs preferences $q_k$, which update the ratings $\rho_k$ in the same way, and absolute scores $s_k\in[0,10]$, along with reasoning. The principles are aggregated into $\bar\rho$ and $\bar s$ by a soft minimum (mean minus half the standard deviation), and all ratings are floored at $0.2$. Further details are in the supplementary.

\paragraph{Numerical Design Principles}
In addition to the VLM evaluation which captures fine-grained semantics and design principles, we measure nine numerical design principles that are geometrically obtained from the generated layouts, each measuring different aspects:
\begin{enumerate*}[label=(\arabic*), itemjoin={{, }}, itemjoin*={{, and }}]
\item \textit{circulation}: the walkable regions and paths to objects
\item \textit{zoning}: arrangement of objects in distinct functional zones
\item \textit{conversation}: arrangement of seating such that they are facing each other and not too far apart
\item \textit{focal point}: the orientation of seating towards a focal object in the scene such as a television or whiteboard
\item \textit{clearance}: free space around objects
\item \textit{wall alignment}: alignment of objects along walls
\item \textit{windows}: windows not blocked by objects
\item \textit{clutter}: how free the floor is of small clutter objects
\item \textit{visual balance}: the spread of objects throughout the scene
\end{enumerate*}.
Each score is measured on a $0$--$1$ scale, with higher being better. We compute $F_{\mathrm{code}}$ as the mean over all applicable scores for a scene. Details are in the supplementary.

\paragraph{Fitness}
The three outputs of the evaluation step are combined into a single fitness score $F$, which is used to select parents, update the archive of layouts, select layouts for refinement, and select the final rearrangements.
\[
 F=\sqrt[3]{F_{\mathrm{rel}}\cdot F_{\mathrm{abs}}\cdot F_{\mathrm{code}}}
\]
$F_{\mathrm{rel}}=\sqrt{\rho_T\,\bar\rho}$ is the combined relative score over prompt adherence and design principles, and $F_{\mathrm{abs}}=\sqrt{s_T\,\bar s}/10$ is the combined absolute score. This geometric aggregation ensures that a layout has to satisfy all three terms.

\subsection{Archive of Diverse Layouts}
\label{sec:archive}
Apart from the mutation strategies and design seeds which explore diverse layouts, we maintain a MAP-Elites archive \cite{mouret2015illuminating} of distinct structural layouts. The cells of the archive are defined by the coarse location of the largest objects in the scene. A new layout replaces an existing layout in a cell only if it has a higher fitness. Additionally, a layout that opens a new archive cell or improves it receives a small probability bonus of being selected as a parent.

\subsection{Application to Rearrangement Planning}
We show an extension of our evolutionary search that continues the existing evolution with an additional move planner evaluation, such that layouts are also steered towards physical feasibility. We execute a fixed move planner that runs during the evaluation step and plans the physical movement from object locations and orientations in the input scene to those in the rearranged layout. Objects are placed on a fine-grained occupancy grid on the floor and moved using a minimum-cost path search over position and orientation (Dijkstra's algorithm). Several different object orders are attempted, and objects that block the path of other objects are temporarily moved to free space. The mutation agent receives the rearrangement plan of its parent as feedback, including a list of objects that are blocked. The fitness for evolution remains as before. The final planned layouts are chosen based on additional movement scores which measure the difficulty of the move normalized by path distance and number of objects. Layouts with high move plan scores are chosen such that their fitness is within $2\%$ of the top fitness candidates. Details are in the supplementary.
\vspace{-0.01cm}
\paragraph{Implementation Details}
We use Claude Opus 5.5 as the mutation and evaluation agent and for generating 4--6 requirements and $K=8$ design seeds. We use Qwen3-VL-32B-Instruct~\cite{bai2025qwen3} for object cards. We use $\lambda=10$ and $\alpha=0.1$. The top-3 layouts by fitness are selected as the output layouts. We use $M=4$ islands and a budget of \$10, which gives on average 35 proposals in 24 minutes.

%% file: sec_arxiv/experiments.tex
\section{Experiments}
\label{sec:experiments}

\subsection{Evaluation setting}
We evaluate \OURS{} and baselines on three datasets: \NumScenesSAGE{} scenes from the original SAGE-10k dataset \cite{xia2026sage}, \NumScenesSAGEOpus{} scenes from SAGE-Opus, an improved version of the SAGE dataset generated using Opus 5.5, and \NumScenesSceneSmith{} scenes from SceneSmith \cite{pfaff2026scenesmith}. Both SAGE and SceneSmith are agentically generated large-scale datasets that include diverse scene types, object layouts and long-tail objects. Scenes are sampled such that the different scene types in each dataset are represented. Input text prompts are either written manually or suggested by Opus 5.5 and edited.

\subsection{Metrics}
We evaluate generated layouts for adherence to the input prompt (Prompt Adherence/PA) and design principles (Design Principles/DP). Since this evaluation requires understanding of complex semantics and spatial reasoning, we use two held-out VLMs which are not used during evolution or in the baselines: Qwen3-VL-32B-Instruct \cite{bai2025qwen3} and Gemma 4 31B Instruct \cite{team2026gemma}. These VLMs are given the generated layouts, and additionally the input text prompt for PA, and prompted for 0--10 scores. We report the mean of the scores output from these two VLMs. We also report the mean of measurable principles which are optimized by our method as Computed Principles (0--1, CP) and selected individual metrics: \textit{circulation} (walkability), \textit{focal} (orientation towards focal object) and \textit{clutter} (absence of small clutter objects on the floor).

We also measure diversity among generated layouts. For a layout $X$ we build a scene graph whose edge set $E_X$ contains an edge $(i,j)$ if objects $i$ and $j$ are closer than $0.6\,$m, $j$ is on top of $i$ or $i$ is close to any wall. $p_X(e)$ is the anchor point of each edge: the midpoint for near edges, and the location of object $i$ for supporting and wall edges. The similarity of two layouts $A$ and $B$ of a room with floor diagonal $d$ is the intersection-over-union of their edge sets weighted by the displacement of the anchor point.

{\small
$$
J(A,B)=\frac{1}{|E_A\cup E_B|}\sum_{e\in E_A\cap E_B}\max\!\Big(0,\;1-\frac{\lVert p_A(e)-p_B(e)\rVert}{d}\Big)
$$
}
 The diversity of a method in a room is the mean pairwise dissimilarity of three of its layouts $X_1,X_2,X_3$,
\[
D=\frac{1}{3}\sum_{1\le a<b\le 3}\big(1-J(X_a,X_b)\big)\in[0,1]
\]
We run each baseline three times and compare with the top-3 proposals from our method by fitness.

\subsection{Baselines}
We compare against \NumBaselines{} state-of-the-art baselines. LayoutGPT \cite{feng2023layoutgpt} generates scenes as stylesheets, LayoutVLM \cite{sun2025layoutvlm}  generates scenes as code with optimizable constraints and ScanEdit \cite{el2025scanedit} uses an LLM agent for problem decomposition followed by constraint optimization. Edit-As-Act \cite{noh2026edit} performs rearrangement by planning the end goal as a set of predicates, and working backwards from the goal to output a minimal sequence of edits. SceneAssistant \cite{luo2026sceneassistant} uses an agentic VLM refinement loop to optimize for physical correctness. LayoutGPT, LayoutVLM and SceneAssistant are adapted to the rearrangement task by conditioning on the input object positions and renders of the scene, while  ScanEdit and Edit-As-Act are designed for scene editing. Additionally, we compare against a fully agentic Opus 5.5 baseline \textit{Opus-Agent} which is given the same input scene and text prompt as ours along with detailed instructions of the rearrangement task and the design principles to be followed. In all cases, we use Opus 5.5 as the LLM/VLM for fair comparison.

\subsection{Comparison to State of the Art}
\input{sec_arxiv/fig_qualitative_sage}
Quantitative results on SAGE, SAGE-Opus and SceneSmith datasets are shown in Tab. \ref{tab:main_combined}. On SAGE, \OURS{} outperforms baselines by \ImprovementPA{}\% on Prompt Adherence, \ImprovementDP{}\% on VLM Design Principles and \ImprovementComputed{}\% on the Computed Principles score. It similarly outperforms baselines on both SAGE-Opus and SceneSmith. Baselines often generate small localized edits, and do not redesign the scene. ScanEdit and SceneAssistant tend to move only objects mentioned in the prompt, while LayoutGPT and LayoutVLM perform larger changes, but make limited use of global scene semantics and hence produce scenes with poorer functionality and aesthetics. Opus-Agent has a stronger prior for design principles and generates plausible scenes, but remains closer to the input scenes.  In contrast, \OURS{} leverages fine-grained object semantic and visual information, detailed feedback from a VLM as well as computed principle scores and optimizes for these by proposing a diverse set of solutions. This is also observed through the scene graph diversity score, where \OURS{} outputs semantically diverse scenes.

\input{sec_arxiv/tables/tab_combined}

A qualitative comparison of the rearranged scenes with the corresponding text prompts is shown in Fig. \ref{fig:qual_sage}. \OURS{} rearranges objects in a balanced manner across the available space, separates objects into functional zones, and ensures symmetry of arrangement while maintaining strong adherence to the input prompt.

\paragraph{Perceptual Study}
We conducted a perceptual study with \UserStudyNumUsers{} participants to compare our generated rearrangements against baselines. Participants are first shown \UserStudyNumPairs{} pairs of scenes from \OURS{} and baselines, and choose the one with higher prompt adherence and design principles. Next, they rate \UserStudyNumSamples{} layouts across all methods on a 1--5 scale for the same two metrics. Results are shown in Tab. \ref{tab:userstudy_combined}. \OURS{} is preferred over all baselines.
\input{sec_arxiv/tables/tab_userstudy_combined}

\subsection{Ablation Studies}
We ablate the different components of our method on a subset of \NumScenesAblations{} SAGE scenes to show their contribution to the full method. Results are shown in Tab. \ref{tab:ablations}. Each component is shown to contribute to the final score either in terms of diversity, prompt adherence or design principles.

\input{sec_arxiv/tables/tab_ablations}

\paragraph{Effect of evolution} \OURS{} generates an evolutionary tree of proposals based on fitness scores. Tab. \ref{tab:ablations} shows that replacing this with one-shot generation of multiple designs leads to lower diversity and aggregate metrics. This is because structurally different layouts are not explored and proposals are not iteratively improved through feedback.
\paragraph{Effect of population and archive} We compare our full method with a variant without the MAP-Elites archive, and with a variant that removes both the population and the archive and instead performs a linear greedy search, always mutating the current best layout, in Tab. \ref{tab:ablations}. Both variants keep our diversity mechanisms in the mutation step (design seeds, exploration strategies and redesigns) and the full evaluation step. These mechanisms preserve diversity, but both variants reduce the quality of the rearrangements in terms of prompt adherence and design principles, whereas our full model achieves both the highest diversity and the highest overall quality.
\paragraph{Effect of diversity strategies} We use only the local exploration strategy by removing the refinement and redesign strategies from the mutation and compare in Tab. \ref{tab:ablations}. This leads to a significantly reduced diversity score of $0.64$ compared to $0.80$ with our full model, and shows that the combination of our multiple diversity strategies leads to better exploration of the design space.
\paragraph{Effect of agentic mutation} We use an agentic mutator that refines a rearrangement proposal in multiple turns before evaluation. We compare this with a single VLM turn mutation in Tab. \ref{tab:ablations}. Using a single turn reduces all the aggregate scores. Our agentic refinement step allows each mutation to fully incorporate VLM feedback and satisfy design principles.
\paragraph{Effect of abstract design principles}
We remove the abstract design principles and the corresponding VLM evaluation during evolution and compare in Tab. \ref{tab:ablations}. This reduces both PA and DP scores, while our full model uses these principles to generate functional, balanced scenes with aesthetic composition.
\paragraph{Effect of numerical design principles}
We remove the computed design principles in the feedback during evolution and compare in Tab. \ref{tab:ablations}. This causes the largest reduction in CP, while \OURS{} uses this feedback to evolve walkable, focused and uncluttered rearrangements among other aspects.

\subsection{Application to Rearrangement Planning}
We evaluate the application of \OURS{} to planning a feasible rearrangement and compare against Opus-Agent given the same feedback from the move planner. The planner tool runs on final layouts of each method to obtain rearrangement plans. We evaluate three metrics:
\begin{enumerate*}[label=(\roman*)]
  \item \textit{blocked}, the fraction of the moved objects that cannot reach their goal;
  \item \textit{extra distance}, the excess distance that objects must be carried, divided by total straight-line distance; and
  \item \textit{walk overhead}, the empty-handed walking distance, divided by carried distance.
\end{enumerate*}
Results are shown in Tab.~\ref{tab:moveplan}. \OURS{} evolves rearrangements with improved feasibility that can be implemented efficiently.
\input{sec_arxiv/tables/tab_pathplanning}

\vspace{0.25cm}
\paragraph{Limitations}

While our method generates high-quality scenes respecting functionality and aesthetics, it leverages VLM feedback, which may be unreliable on very small objects. Our method operates on indoor scenes containing rigid objects and does not consider potential articulations or deformations of objects.

%% file: sec_arxiv/fig_qualitative_sage.tex
\begin{figure*}[p]
\centering
\setlength{\tabcolsep}{1pt}
\renewcommand{\arraystretch}{0.6}
\begin{tabular}{ccccc}
 & \parbox[b]{0.2115\textwidth}{\centering\scriptsize\itshape ``A restaurant setup for large groups of 8 people each.''} & \parbox[b]{0.2342\textwidth}{\centering\scriptsize\itshape ``A themed pub with one drink per person on the table and the chairs arranged around the table''} & \parbox[b]{0.1765\textwidth}{\centering\scriptsize\itshape ``A focused home office for video calls: the desk facing into the room with a calm background behind it, good light from the floor lamps, and the bookshelf or other storage within reach of the chair.''} & \parbox[b]{0.2204\textwidth}{\centering\scriptsize\itshape ``A hotel lobby turned into a conference check-in area: the reception desk facing the entrance as the registration counter. The seating is arranged on the sides and the luggage cart is stowed away.''} \\[2pt]
\rotatebox[origin=c]{90}{\parbox{0.13\textwidth}{\centering\scriptsize Input}} & \raisebox{-.5\height}{\includegraphics[width=0.2115\textwidth]{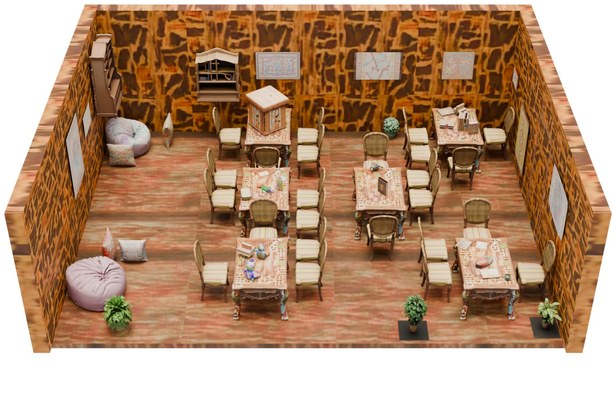}} & \raisebox{-.5\height}{\includegraphics[width=0.2342\textwidth]{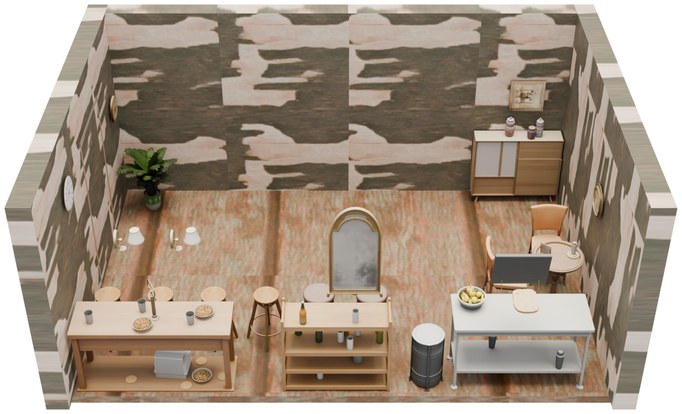}} & \raisebox{-.5\height}{\includegraphics[width=0.1765\textwidth]{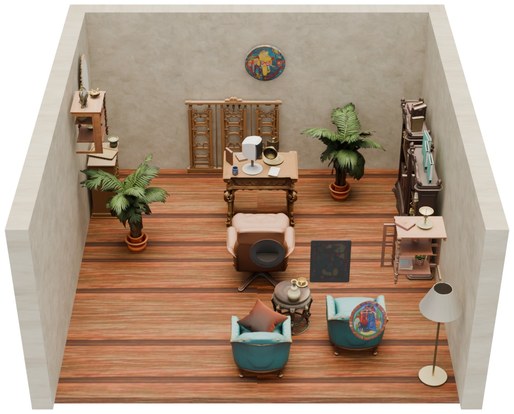}} & \raisebox{-.5\height}{\includegraphics[width=0.2204\textwidth]{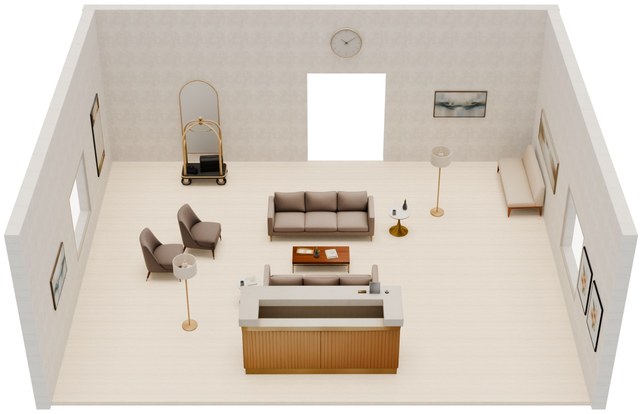}} \\
\rotatebox[origin=c]{90}{\parbox{0.13\textwidth}{\centering\scriptsize LayoutGPT}} & \raisebox{-.5\height}{\includegraphics[width=0.2115\textwidth]{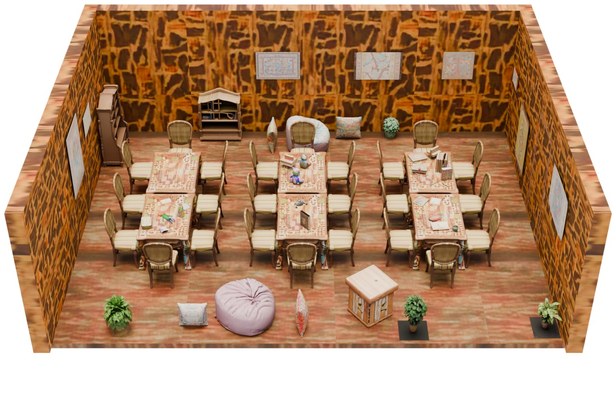}} & \raisebox{-.5\height}{\includegraphics[width=0.2342\textwidth]{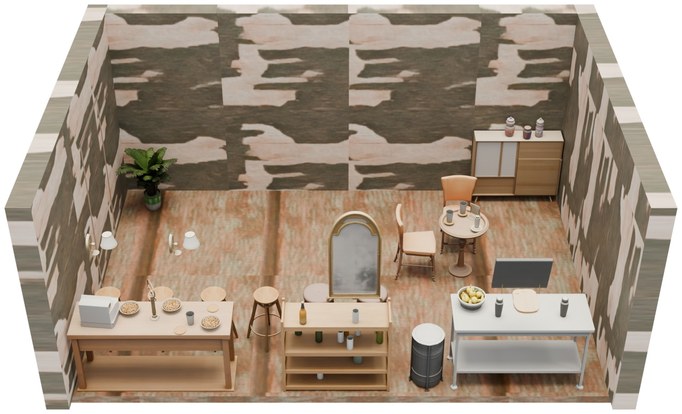}} & \raisebox{-.5\height}{\includegraphics[width=0.1765\textwidth]{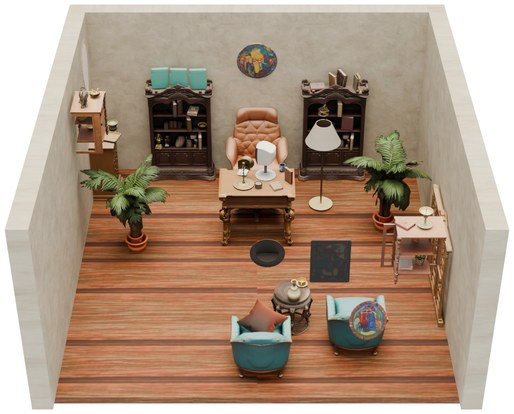}} & \raisebox{-.5\height}{\includegraphics[width=0.2204\textwidth]{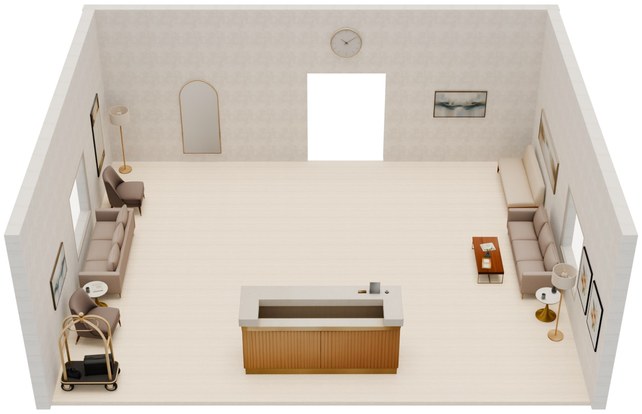}} \\
\rotatebox[origin=c]{90}{\parbox{0.13\textwidth}{\centering\scriptsize LayoutVLM}} & \raisebox{-.5\height}{\includegraphics[width=0.2115\textwidth]{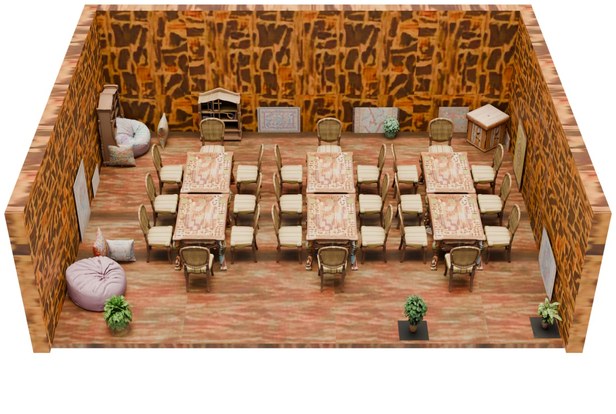}} & \raisebox{-.5\height}{\includegraphics[width=0.2342\textwidth]{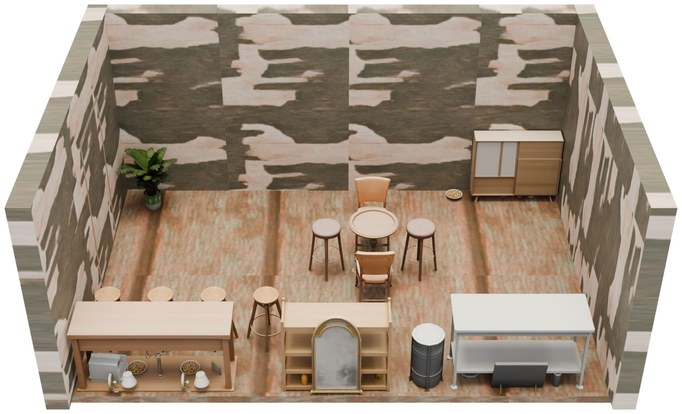}} & \raisebox{-.5\height}{\includegraphics[width=0.1765\textwidth]{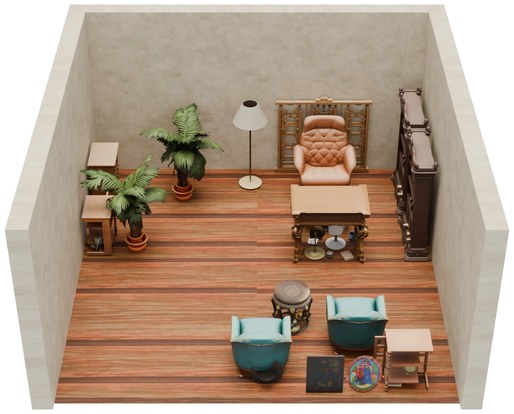}} & \raisebox{-.5\height}{\includegraphics[width=0.2204\textwidth]{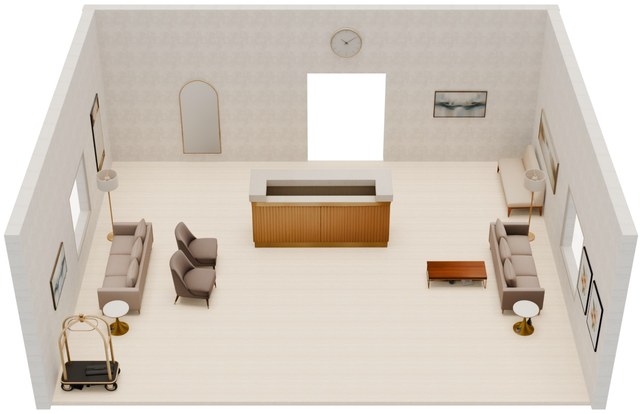}} \\
\rotatebox[origin=c]{90}{\parbox{0.13\textwidth}{\centering\scriptsize ScanEdit}} & \raisebox{-.5\height}{\includegraphics[width=0.2115\textwidth]{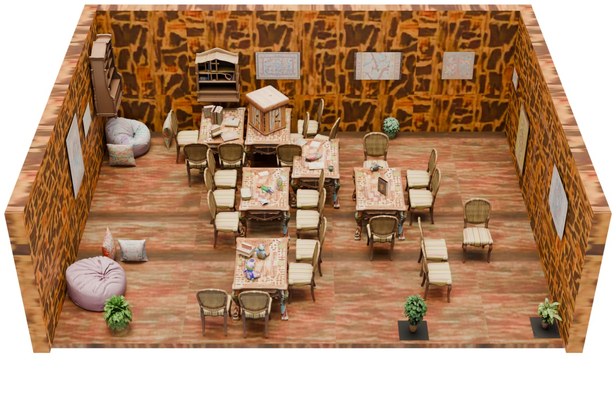}} & \raisebox{-.5\height}{\includegraphics[width=0.2342\textwidth]{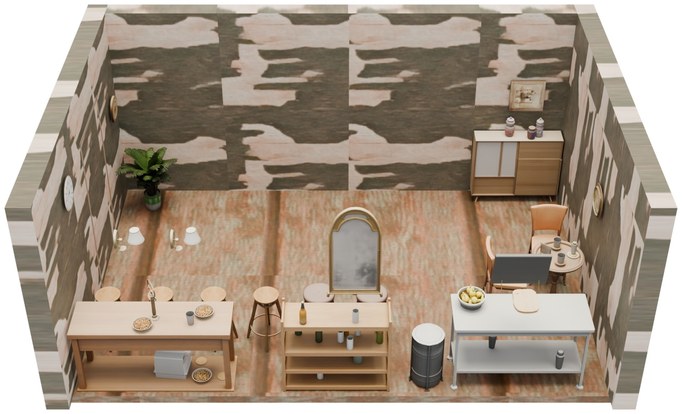}} & \raisebox{-.5\height}{\includegraphics[width=0.1765\textwidth]{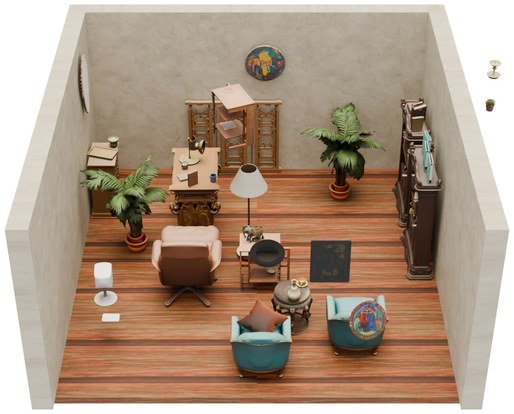}} & \raisebox{-.5\height}{\includegraphics[width=0.2204\textwidth]{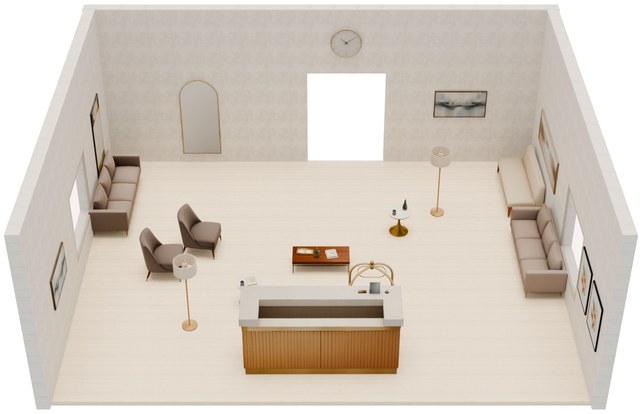}} \\
\rotatebox[origin=c]{90}{\parbox{0.13\textwidth}{\centering\scriptsize Edit-As-Act}} & \raisebox{-.5\height}{\includegraphics[width=0.2115\textwidth]{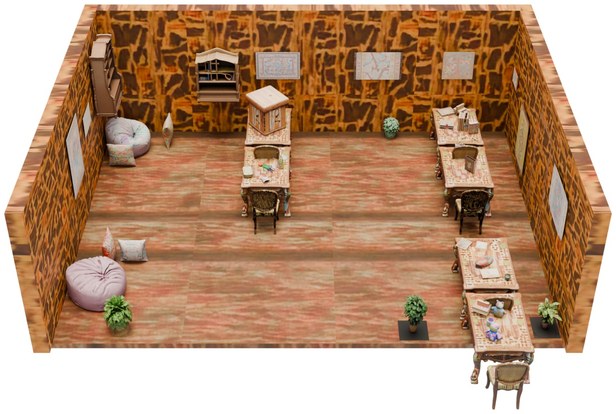}} & \raisebox{-.5\height}{\includegraphics[width=0.2342\textwidth]{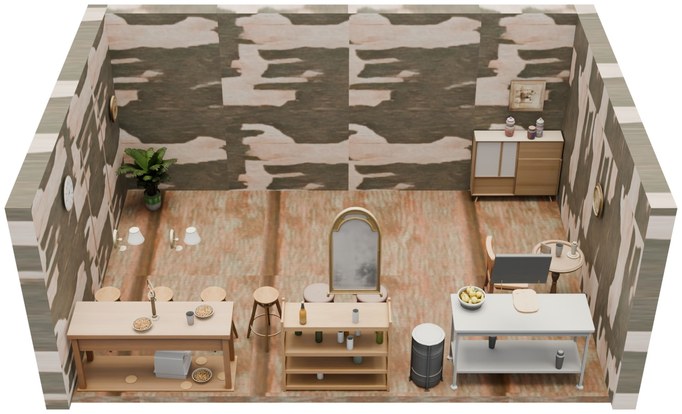}} & \raisebox{-.5\height}{\includegraphics[width=0.1765\textwidth]{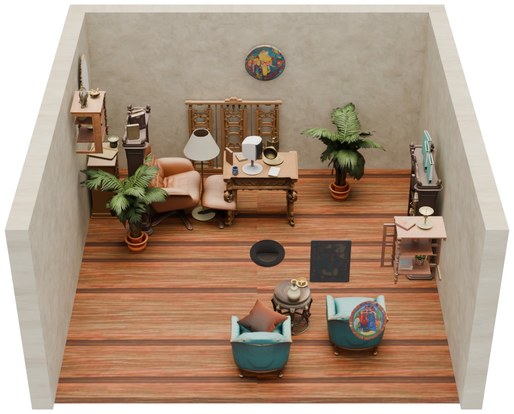}} & \raisebox{-.5\height}{\includegraphics[width=0.2204\textwidth]{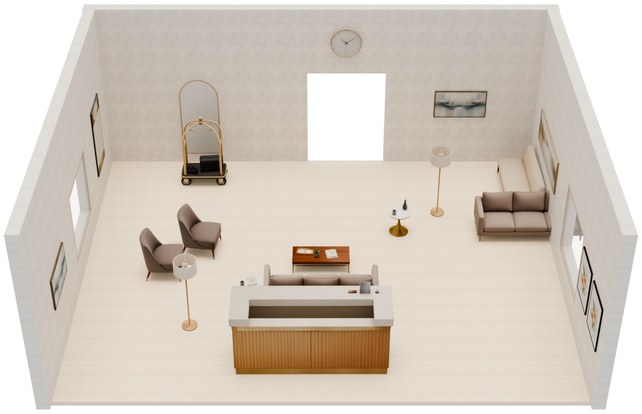}} \\
\rotatebox[origin=c]{90}{\parbox{0.13\textwidth}{\centering\scriptsize SceneAssistant}} & \raisebox{-.5\height}{\includegraphics[width=0.2115\textwidth]{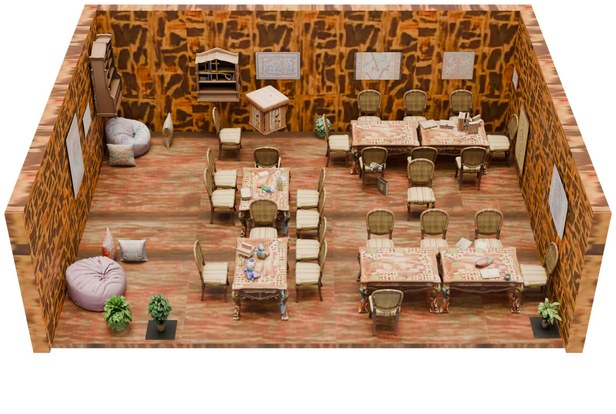}} & \raisebox{-.5\height}{\includegraphics[width=0.2342\textwidth]{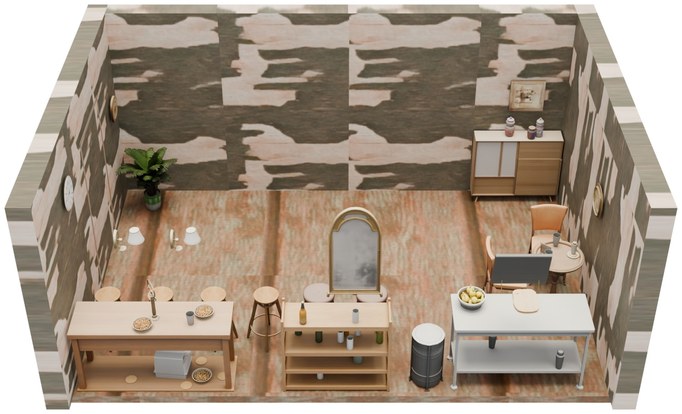}} & \raisebox{-.5\height}{\includegraphics[width=0.1765\textwidth]{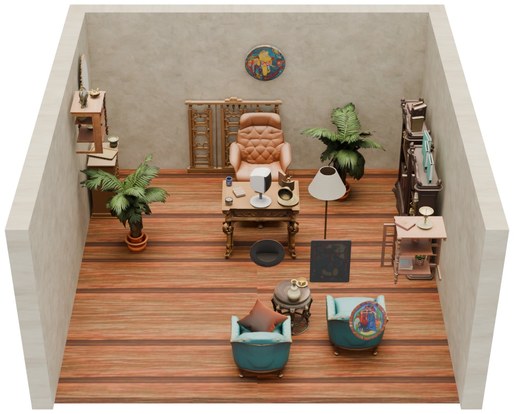}} & \raisebox{-.5\height}{\includegraphics[width=0.2204\textwidth]{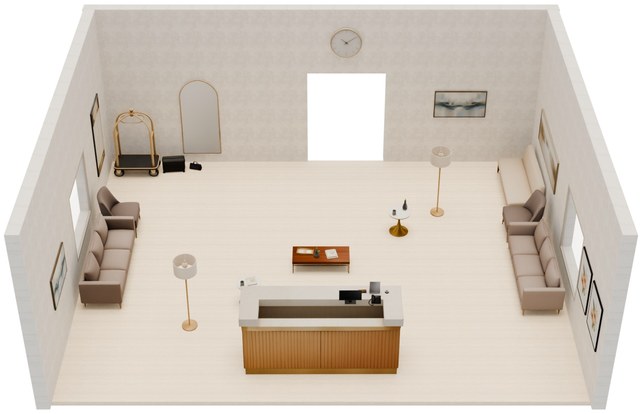}} \\
\rotatebox[origin=c]{90}{\parbox{0.13\textwidth}{\centering\scriptsize Opus-Agent}} & \raisebox{-.5\height}{\includegraphics[width=0.2115\textwidth]{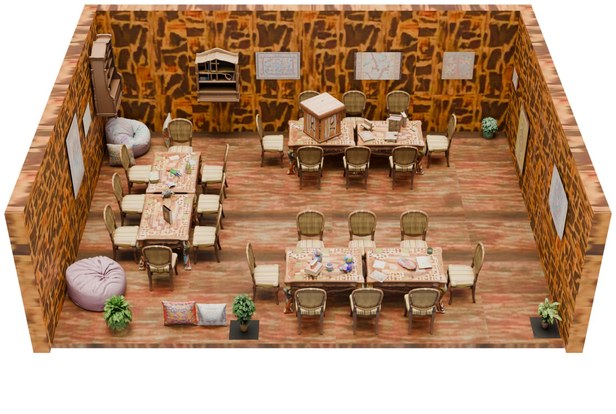}} & \raisebox{-.5\height}{\includegraphics[width=0.2342\textwidth]{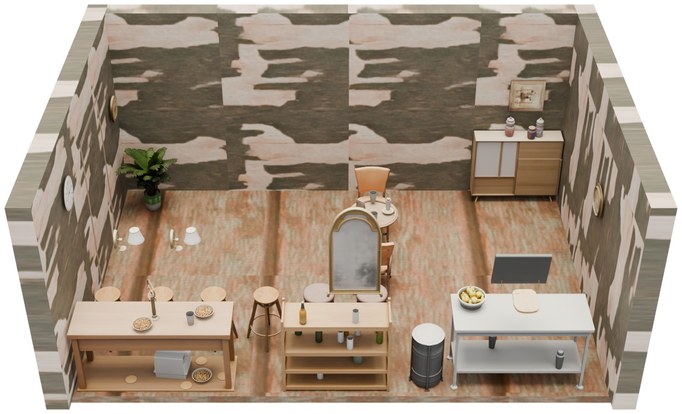}} & \raisebox{-.5\height}{\includegraphics[width=0.1765\textwidth]{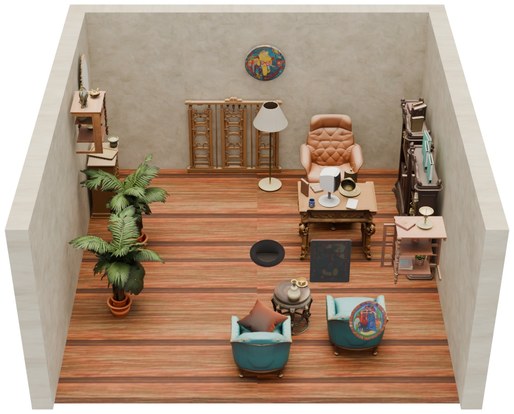}} & \raisebox{-.5\height}{\includegraphics[width=0.2204\textwidth]{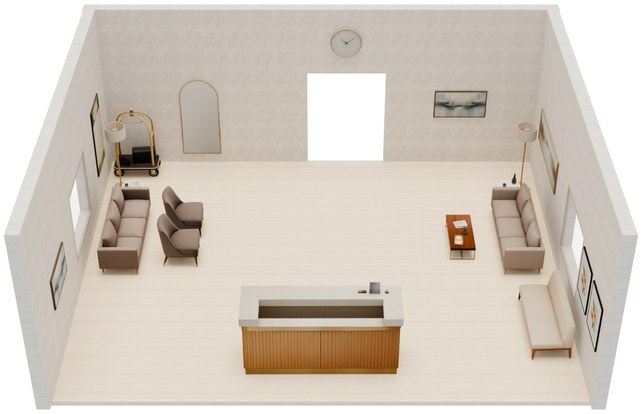}} \\
\rotatebox[origin=c]{90}{\parbox{0.13\textwidth}{\centering\scriptsize \OURS{}}} & \raisebox{-.5\height}{\includegraphics[width=0.2115\textwidth]{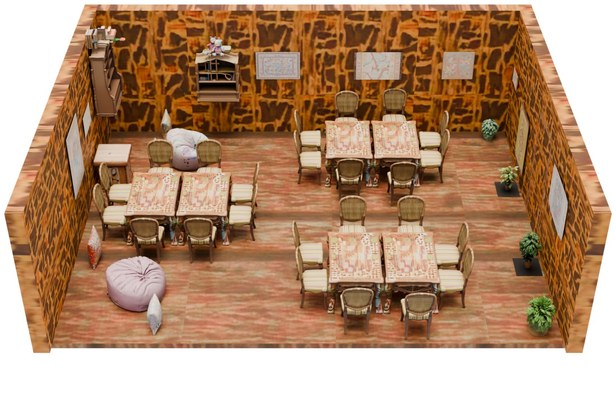}} & \raisebox{-.5\height}{\includegraphics[width=0.2342\textwidth]{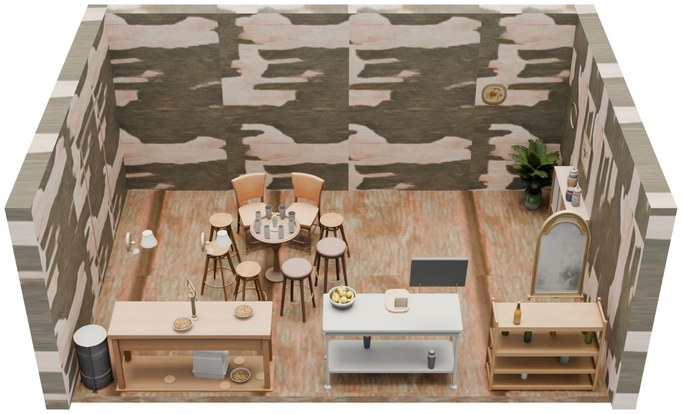}} & \raisebox{-.5\height}{\includegraphics[width=0.1765\textwidth]{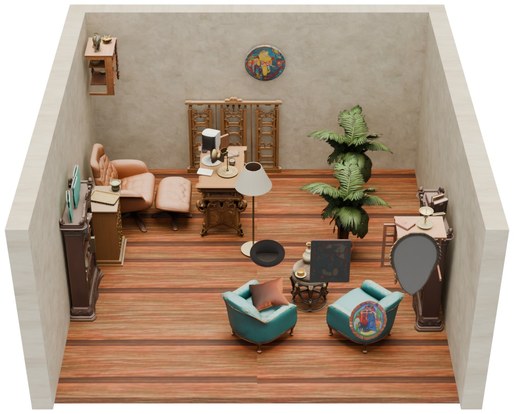}} & \raisebox{-.5\height}{\includegraphics[width=0.2204\textwidth]{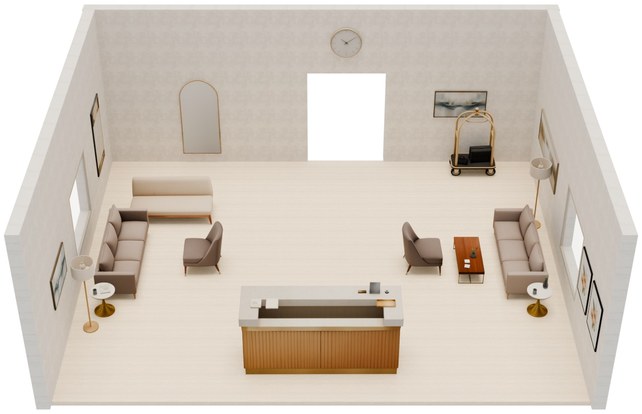}} \\
\end{tabular}
\caption{Qualitative comparison of \OURS{} against baselines on the SAGE, SAGE-Opus and SceneSmith datasets. The input scenes and text prompts are shown on the top. \OURS{} generates rearrangements that strongly satisfy the input prompt as well as common interior design principles such as reachability of objects, functional zoning and visual balance.}
\label{fig:qual_sage}
\end{figure*}

%% file: sec_arxiv/tables/tab_combined.tex
\begin{table}[t]
\centering
\footnotesize
\setlength{\tabcolsep}{3pt}
\begin{tabular}{clccccccc}
\toprule
 & Method & PA $\uparrow$ & DP $\uparrow$ & circ $\uparrow$ & foc $\uparrow$ & clut $\uparrow$ & CP $\uparrow$ & Div. $\uparrow$ \\
\midrule
\multirow{7}{*}{\rotatebox[origin=c]{90}{SAGE}} & LayoutGPT & 6.20 & 7.06 & 0.57 & 0.64 & \underline{0.56} & 0.64 & \underline{0.39} \\
 & LayoutVLM & \underline{6.32} & 6.73 & 0.50 & 0.54 & 0.54 & 0.63 & 0.36 \\
 & ScanEdit & 4.94 & 6.44 & 0.44 & \underline{0.67} & 0.55 & 0.61 & 0.24 \\
 & Edit-As-Act & 4.39 & 5.98 & 0.56 & 0.63 & 0.53 & 0.60 & 0.23 \\
 & SceneAssistant & 4.94 & 6.64 & 0.55 & 0.63 & 0.55 & 0.64 & 0.07 \\
 & Opus-Agent & 6.06 & \underline{7.16} & \underline{0.62} & 0.64 & 0.55 & \underline{0.66} & 0.23 \\
\cmidrule(l){2-9}
 & \OURS{} & \textbf{6.64} & \textbf{7.31} & \textbf{0.81} & \textbf{0.96} & \textbf{0.65} & \textbf{0.76} & \textbf{0.80} \\
\midrule
\multirow{7}{*}{\rotatebox[origin=c]{90}{SAGE-Opus}} & LayoutGPT & 6.16 & 7.31 & 0.68 & 0.65 & 0.73 & 0.70 & 0.27 \\
 & LayoutVLM & \underline{6.34} & 6.69 & 0.62 & 0.65 & 0.54 & 0.66 & \underline{0.34} \\
 & ScanEdit & 5.11 & 6.72 & 0.49 & \underline{0.69} & \underline{0.76} & 0.66 & 0.25 \\
 & Edit-As-Act & 4.53 & 6.43 & 0.63 & 0.56 & 0.73 & 0.68 & 0.19 \\
 & SceneAssistant & 5.44 & 7.15 & 0.63 & 0.65 & 0.73 & 0.67 & 0.06 \\
 & Opus-Agent & \underline{6.34} & \underline{7.34} & \underline{0.70} & 0.66 & 0.74 & \underline{0.71} & 0.18 \\
\cmidrule(l){2-9}
 & \OURS{} & \textbf{6.55} & \textbf{7.46} & \textbf{0.78} & \textbf{0.94} & \textbf{0.81} & \textbf{0.78} & \textbf{0.77} \\
\midrule
\multirow{7}{*}{\rotatebox[origin=c]{90}{SceneSmith}} & LayoutGPT & \underline{7.32} & 7.75 & 0.93 & 0.42 & 0.76 & 0.66 & 0.22 \\
 & LayoutVLM & 7.08 & 7.54 & 0.93 & 0.43 & 0.72 & 0.65 & \underline{0.29} \\
 & ScanEdit & 5.54 & 7.08 & 0.90 & 0.42 & 0.74 & 0.66 & 0.20 \\
 & Edit-As-Act & 5.04 & 6.93 & 0.86 & 0.59 & 0.76 & 0.64 & 0.17 \\
 & SceneAssistant & 6.74 & 7.60 & \underline{0.94} & 0.56 & \underline{0.77} & \underline{0.67} & 0.14 \\
 & Opus-Agent & 7.06 & \underline{7.78} & 0.90 & \underline{0.76} & 0.76 & 0.66 & 0.28 \\
\cmidrule(l){2-9}
 & \OURS{} & \textbf{7.48} & \textbf{7.90} & \textbf{0.98} & \textbf{0.92} & \textbf{0.80} & \textbf{0.74} & \textbf{0.74} \\
\bottomrule
\end{tabular}%
\caption{Quantitative comparison on the SAGE, SAGE-Opus, and SceneSmith datasets. PA: prompt adherence (VLM judges, 0--10); DP: design principles (VLM judges, 0--10); CP: computed design principles (mean); Div: Scene-graph diversity. The best results per dataset are \textbf{bolded} and the second best are \underline{underlined}.}
\label{tab:main_combined}
\end{table}

%% file: sec_arxiv/tables/tab_userstudy_combined.tex
\begin{table}[t]
\centering
\footnotesize
\setlength{\tabcolsep}{3pt}
\newcolumntype{C}{>{\centering\arraybackslash}p{1.2cm}}
\begin{tabular}{lCC|CC}
\toprule
 & \multicolumn{2}{c|}{Preference study} & \multicolumn{2}{c}{\makecell{Scores (1--5) $\uparrow$}} \\
\cmidrule(lr){2-3} \cmidrule(lr){4-5}
Method & PA & DP & PA & DP \\
\midrule
LayoutGPT & 88.9 & 91.1 & 2.69 & 2.58 \\
LayoutVLM & 88.9 & 93.3 & \underline{3.60} & \underline{3.60} \\
ScanEdit & 95.6 & 84.4 & 2.24 & 2.24 \\
Edit-As-Act & 91.1 & 95.6 & 1.76 & 1.78 \\
SceneAssistant & 80.0 & 77.8 & 3.00 & 2.76 \\
Opus-Agent & 95.6 & 88.9 & 3.38 & 3.29 \\
\midrule
\OURS{} & -- & -- & \textbf{4.29} & \textbf{4.27} \\
\bottomrule
\end{tabular}%
\caption{User study on the SAGE, SAGE-Opus, and SceneSmith datasets. Left: share (\%) of pairwise choices preferring \OURS{} over each baseline. Right: mean rating (1--5) of each method. PA: prompt adherence, DP: design principles. }
\label{tab:userstudy_combined}
\end{table}

%% file: sec_arxiv/tables/tab_ablations.tex
\begin{table}[t]
\centering
\footnotesize
\setlength{\tabcolsep}{3pt}
\resizebox{\columnwidth}{!}{%
\begin{tabular}{lccccccc}
\toprule
Method & PA $\uparrow$ & DP $\uparrow$ & circ $\uparrow$ & foc $\uparrow$ & clut $\uparrow$ & CP $\uparrow$ & Div. $\uparrow$ \\
\midrule
w/o evolution & \underline{7.23} & 7.45 & 0.67 & \textbf{1.00} & 0.55 & 0.69 & 0.61 \\
w/o population \& archive & \underline{7.23} & 7.49 & 0.72 & 0.92 & \textbf{0.66} & \underline{0.73} & 0.75 \\
w/o MAP-Elites archive & 7.00 & 7.44 & 0.68 & \underline{0.94} & 0.61 & \underline{0.73} & \underline{0.79} \\
w/o refine + redesign & \textbf{7.37} & 7.41 & 0.71 & 0.92 & 0.58 & 0.71 & 0.64 \\
w/o agentic mutation & 7.10 & 7.48 & 0.70 & 0.93 & 0.61 & 0.70 & 0.68 \\
w/o abstract principles & \underline{7.23} & \underline{7.55} & \textbf{0.75} & 0.87 & 0.56 & 0.72 & 0.75 \\
w/o numeric principles & 7.13 & 7.41 & 0.62 & 0.61 & 0.51 & 0.65 & 0.71 \\
\midrule
Full model \textit{(Ours)} & \textbf{7.37} & \textbf{7.65} & \underline{0.74} & 0.91 & \underline{0.64} & \textbf{0.75} & \textbf{0.80} \\
\bottomrule
\end{tabular}%
}
\caption{Ablation study of the different components of our method on a subset of 15 SAGE scenes. Each component is ablated independently from the full model.}
\label{tab:ablations}
\end{table}

%% file: sec_arxiv/tables/tab_pathplanning.tex
\begin{table}[t]
\centering
\footnotesize
\setlength{\tabcolsep}{3pt}
\resizebox{\columnwidth}{!}{%
\begin{tabular}{lccc}
\toprule
Method & blocked $\downarrow$ & extra distance $\downarrow$ & walk overhead $\downarrow$ \\
\midrule
Opus-Agent & 0.022 & 0.39 & 0.57 \\
\OURS{} & \textbf{0.019} & \textbf{0.30} & \textbf{0.48} \\
\bottomrule
\end{tabular}%
}
\caption{Rearrangement planning metrics on 33 scenes from the SAGE dataset. \OURS{} produces more feasible rearrangements.}
\label{tab:moveplan}
\end{table}

%% file: sec_arxiv/conclusion.tex
\section{Conclusion}
\label{sec:conclusion}
We presented \OURS{}, an agentic evolutionary search method for 3D scene rearrangement. We showed that VLM mutation and evaluation produce diverse, functional and aesthetic rearrangements, and hope this will encourage future work on evolutionary methods for 3D generation.

\section*{Acknowledgements}
This project was supported by the ERC Starting Grant SpatialSem (101076253). The research presented is supported by the TUM Georg Nemetschek Institute Artificial Intelligence for the Built World.

\section*{LLM Usage Statement}
LLMs (Claude Opus 5.5 and older versions) were used in the development of this work. Opus was accessed through  Claude Code and the web interface. It was used for literature survey, ideation, code development and proofreading the paper. The text in the paper is primarily written by hand.

%% file: sec_arxiv/suppl.tex
\clearpage
\setcounter{page}{1}
\maketitlesupplementary

\section{Additional Results and Analysis}
\subsection{Ablations}
Additional ablation studies are shown in Tab. \ref{tab:suppl_ablations}.
\input{sec_arxiv/suppl_tables/ablation}

\paragraph{Other VLMs} We compare against three open-weight models which replace Opus 5.5 in our mutation step: Qwen-3.8-27B \cite{qwen38}, Muse-Glimmer-30B \cite{abtahizadeh2026muse} and MiniCPM-4.7 \cite{cui2026minicpmo45realtimefullduplex}. We also compare against smaller paid models: Claude Haiku 5.5 and Claude Sonnet 5.5. \OURS{} has slightly reduced performance with smaller open-weight VLMs, while being able to leverage a frontier VLMs such as Opus 5.5 and Sonnet 5.5 for increased prompt adherence and design principle scores.

\subsection{VLM Judge Reproducibility}
\input{sec_arxiv/tables/tab_vlm_variance}
We repeat the VLM evaluation three times on the SAGE ablation set to measure its reproducibility. Results are shown in Tab. \ref{tab:vlm_variance}. The VLM scores show only small variations across runs, and \OURS{} consistently outperforms the Opus-Agent baseline.

\subsection{Inter-VLM Agreement}
\input{sec_arxiv/tables/tab_vlm_agreement}
\input{sec_arxiv/fig_vlm_agreement}
We compare the scores output by the two held-out VLM judges: Qwen3-VL-32B-Instruct and Gemma 4 31B Instruct in Fig. \ref{fig:vlm_agreement}. We see that both VLMs have the same trends across all three datasets and both prompt adherence and design principles. Differences in the absolute scores are expected since the VLMs are trained on different data distributions. \OURS{} largely matches or outperforms baselines when evaluated by either VLM independently.

We also report the Kendall's $\tau$, which measures how similarly the two VLMs rank the seven methods. $\tau$ ranges from $-1$ (inverse ordering) to $1$ (identical ordering). $\tau > 0.7$ is generally considered to show strong agreement. We obtain $\tau$ between $0.71$ and $0.90$ for both metrics on all three datasets, as shown in Tab.~\ref{tab:vlm_agreement}.  Although Gemma 4 scores more strictly (lower) than Qwen3-VL, both judges agree on which methods produce better rearrangements in terms of prompt adherence and design principles.

\subsection{Qualitative Results}
\input{sec_arxiv/fig_qualitative_suppl}
We show additional qualitative comparisons of \OURS{} and baselines in Fig. \ref{fig:qual_suppl}. \OURS{} generates rearrangements that closely adhere to the input text prompt and follow common design principles, making balanced use of space with all objects reachable, resulting in aesthetically pleasing scenes.

\subsection{Diversity of Generated Scenes}
\input{sec_arxiv/fig_diversity_suppl}
\OURS{} generates diverse rearrangements for a given input scene and text prompt, shown in Fig.  \ref{fig:div_suppl}.

\section{Implementation Details}
\subsection{Evolution}
The evolved program is a single \texttt{rearrange} function that takes the input scene and text prompt as input, and returns the rearranged object locations and orientations. It has access to an API with useful functions such as \texttt{move}, \texttt{free\_spot}, \texttt{place\_near} and \texttt{push\_against\_wall}.
We use a modified version of ShinkaEvolve~\cite{lange2026shinkaevolve}, and perform full rewrites of the rearrangement program. We sample inspiration programs across all islands.

\paragraph{Design Seeds} The evolution starts with a single LLM call that outputs $K=8$ short design ideas, which are then used for the first $K$ mutations.

\subsection{MAP-Elites Archive}
Cells in the archive are created according to the floor quadrant occupied by each anchor object. Anchor objects are the two largest movable objects on the floor, followed by any large objects that are mentioned in the prompt ($\geq0.4\,$m$^2$) and other large objects ($\geq0.6\,$m$^2$). When a new layout opens a new archive cell, it receives a fitness bonus of $0.03$ for parent sampling. If a layout improves an existing cell, it receives a bonus of $0.01$.

\paragraph{Agentic Mutation}
The mutation mode is selected based on the current state of the evolution and the selected parent, from among three options: exploration (with several sub-options), refinement (of existing candidates) and redesign (large changes from scratch).

Refinement involves smaller changes to an existing high-quality layout. When the parent is one of the best three layouts by fitness, it is refined.

Redesigns are scheduled at fixed exponentially increasing intervals (1st, 3rd and 7th evaluation after the seed phase ($2^j-1$)), and each followed by a fixed number of refinements. A redesign mutation ignores the input object positions and earlier layouts, and follows the least-used design seed to arrange objects freely. It must move at least 70\% of the movable floor objects. Redesign-refinements must preserve the new structure of the redesign.

For every other selection of parent, the mutator explores new designs. It has a fixed set of exploration strategies and can access statistics on how often each strategy has been tried and how often it produced a new best layout or a new archive cell. If a strategy failed twice on the same parent, it cannot be selected in future mutations.

Mutation is performed by the Claude Opus 5.5 VLM with file and shell tools in a sandboxed environment. Each mode has different reasoning and turn budgets:
\begin{enumerate*}[label=(\arabic*), itemjoin={{, }}, itemjoin*={{, or }}]
    \item seed: low effort, 12 turns
    \item exploration and refinement: low effort, no thinking, 8 turns
    \item redesign: medium effort, thinking, 10 turns
    \item redesign-refinement: medium effort, thinking, 8 turns.
\end{enumerate*}

\subsection{Validity Check}
A rearrangement is invalid if:
\begin{enumerate*}[label=(\arabic*), itemjoin={{, }}, itemjoin*={{, or }}]
\item it moves a fixed object
\item it moves an object outside the scene bounds
\item it causes two object bounding boxes to overlap by more than $0.006m^3$ beyond their overlap in the input, or their mesh overlap (1\,cm voxels) grows by more than $0.0005m^3$
\item the center of a supported object leaves the support
\item it places an object taller than $0.25\,$m in front of a door
\item it places an object taller than $1.3\,$m within $0.4\,$m of a window
\item it disconnects doors that were originally connected in the input
\end{enumerate*}.
Objects on top of other objects move along with the supporting object. Invalid layouts do not proceed to the evaluation step, instead they receive a low fitness score.

\subsection{Abstract Design Principles}
We use ten common abstract design principles, which are given as context to the VLM mutation and evaluation (judge) agents.

\begin{itemize}[leftmargin=*,itemsep=1pt]
\item \textit{Zoning}: furniture must be grouped into zones such as seating, dining, work, sleep, reading, and separated by walking paths between them. Each zone should preferably have its own lighting.
\item \textit{Floating and anchoring}: seating and furniture must not always be lined against the walls, they should occupy central positions in the room, preferably anchored by a rug on the floor.
\item \textit{Architecture}: the largest or dominant piece of furniture such as a sofa or bed must be aligned with the room's strongest architectural feature, such as a large window, wall or fireplace. Furniture is placed parallel or perpendicular to walls and to other furniture.
\item \textit{Balance and proportion}: visual weight (large/small furniture) must be distributed evenly across the scene. In a single-purpose room, the main furniture must use about two-thirds of the space, so that it is not overcrowded and does not leave large spaces unused.
\item \textit{Emphasis}: a room must preferably have one clear focal point, such as a TV against a wall, a large window, whiteboard or fireplace. The main seating can be oriented towards it symmetrically.
\item \textit{Composition}: objects must be arranged through deliberate symmetry, radially or asymmetrically. Objects can be repeated at regular intervals through spacing or orientation.
\item \textit{Comfort}: the main seating, beds and desks can be placed along walls such that they have a view of the room and window or door diagonally.
\item \textit{Harmony and contrast}: objects with matching or complementary styles are grouped and placed together, contrast can be used sparingly and deliberately.
\item \textit{Function}: the room must have clear walking paths (main paths $\geq91\,$cm, secondary $\geq76\,$cm), space around objects (\eg $76\,$cm in front of seats, $61\,$cm in front of cabinets, $91\,$cm around a dining table), usable distances between related objects (\eg coffee table 35--45\,cm in front of a sofa, seats for conversation must be 1.2--2.4\,m apart). Doors must not be blocked.
\item \textit{Order}: every object must be deliberately placed with a reason. Small objects must be gathered onto surfaces and not be left on the floor unless required.
\end{itemize}

\subsection{Computed Principles and Metrics}
The geometrically computed principles give numerical scores independent of the VLMs and are used as feedback as well as in final evaluation. The scores are in the range $[0,1]$ and the mean is taken over all scores that apply to a scene.

\paragraph{Circulation} The path to seating and storage objects must be at least $0.6$m, the circulation score averaged over all such objects is $\mathrm{clip}((w-0.6)/0.3,0,1)$ where $w$ is the path width in metres
\paragraph{Zoning} Zones are  created based on large functional objects (sleep, dining, seating, work), and nearby objects are grouped into these zones. The score is the average of fraction of furniture that belongs to a zone, the floor coverage of the zones and separation of zones.
\paragraph{Conversation} Each seat should be angled towards the group center, the score is 1 within $45^\circ$ and decreases to 0 at $135^\circ$. Seats should be at most $3.2\,$m apart.
\paragraph{Focal point} The largest seating object should face a focal object such as a TV or whiteboard, the score is 1 within $30^\circ$ and decreases to 0 at $90^\circ$.
\paragraph{Clearance} Seats are placed at most $0.6\,$m away from a table or low storage surface and $0.3$--$0.5\,$m from coffee tables. A chair at a table must have $0.75\,$m of free space behind it and a storage unit must have $0.8\,$m in front of it. Each check scores 1 when satisfied and reduces linearly otherwise.
\paragraph{Wall alignment} Similar to clearance, tall objects ($>1.2\,$m) must be within $0.15\,$m of a wall, and furniture must be perpendicular or parallel to the walls or to a neighbour within $5^\circ$. Seats must not face walls.
\paragraph{Windows} Objects taller than $1\,$m do not stand within $0.6\,$m of a window.
\paragraph{Clutter} Small floor objects should be placed near a wall or other furniture. They should have a density below $0.3$ per m$^2$. A table top should carry at most 6 items per m$^2$.
\paragraph{Visual balance} The visual weight of an object (footprint $\times$ height) must be split evenly along both room axes. Tall objects must be spread along all the walls. The main furniture group must be mirror-symmetric about its center, such that objects are within $0.35\,$m of the mirrored position. The largest empty square on the floor must be at most 20\% of the floor.

\subsection{VLM Evaluation and Fitness}
\paragraph{Fitness judge} We use Claude Opus 5.5 as the VLM judge for candidate layouts to evaluate prompt adherence and design principles. It is called without any tools. It receives two $1280\times720$ renders each of the parent and new candidate from two fixed viewpoints, as well as a labeled top-down plan showing object bounding boxes and a list of object positions. It outputs relative preference and absolute scores for the prompt requirements and the design principles, along with short reasoning which forms the feedback to the next mutations.

\paragraph{Final evaluation} We use two held-out VLM judges (Qwen3-VL-32B-Instruct and Gemma 4 31B Instruct) which receive top-down renders of the input and output layouts, two oblique views of the output, close-ups of up to four large moved objects and the labeled plan. Scores are averaged over the two judges and across all evaluated scenes.

\subsection{Rearrangement Planning}
We resume each evolution run with an additional budget of \$5. A fixed move planner is run on every valid mutation output (candidate layout) during evaluation. The list of blocked objects, object parking moves and detours is added to the feedback so that the next mutation can resolve them.

Small supported items such as books on a table are moved along with the supporting item. The order of moving objects is chosen by attempting several move heuristic orders (largest object first, nearest object next, object dependency order). The movement is planned using a $5$\,cm grid and $15^\circ$ orientation bins. Finally, we select the plan with the highest move score among the layouts within $2\%$ of the best fitness.

The move plan is evaluated based on the difficulty of the move, normalized by the number of objects moved and the path distance. This allows methods to perform larger rearrangements and achieve diversity without being penalized for it. The computed scores during evolution are:
\begin{enumerate*}[label=(\arabic*), itemjoin={{, }}, itemjoin*={{, and }}]
\item \textit{detour}: the ratio of moved path length $\ell$ to the straight-line distance $d$
\item \textit{hardest move}: the largest ratio $\ell/d$
\item \textit{re-handling}: the number of handling moves (moving it to its goal or to a temporary spot) per object
\item \textit{dependency}: the share of rearranged objects that cannot move straight to their goal in the input scene
\item \textit{walk overhead}: ratio of empty-handed walking distance between moves to the carried path length
\end{enumerate*}.
Each score is normalized to $[0,1]$ using the 10th and 90th percentile over all layouts, such that 1 is the best score. The move score is their mean divided by $1+n_b$, where $n_b$ is the number of blocked objects.

%% file: sec_arxiv/suppl_tables/ablation.tex
\begin{table}[h]
\centering
\setlength{\tabcolsep}{3pt}
\resizebox{\columnwidth}{!}{%
\begin{tabular}{lccccccc}
\toprule
Method & PA $\uparrow$ & DP $\uparrow$ & circ $\uparrow$ & foc $\uparrow$ & clut $\uparrow$ & CP $\uparrow$ & Div. $\uparrow$ \\
\midrule
Ours (Qwen3.8-27B) & 6.23 & 6.86 & 0.54 & 0.76 & 0.50 & 0.64 & 0.72 \\
Ours (Muse-Glimmer-30B) & 5.93 & 6.69 & 0.61 & 0.80 & \underline{0.56} & 0.65 & 0.67 \\
Ours (MiniCPM-V-4.7-35B-A3B) & 6.73 & 7.08 & \underline{0.67} & \underline{0.92} & 0.54 & 0.67 & 0.70 \\
\midrule
Ours (Haiku 5.5) & 6.70 & 6.99 & 0.60 & 0.89 & 0.51 & 0.67 & 0.66 \\
Ours (Sonnet 5.5) & \underline{7.20} & \underline{7.31} & 0.60 & \textbf{0.99} & 0.52 & \underline{0.69} & \underline{0.73} \\
\midrule
Full model \textit{(Ours)} & \textbf{7.37} & \textbf{7.65} & \textbf{0.74} & 0.90 & \textbf{0.64} & \textbf{0.75} & \textbf{0.80} \\
\bottomrule
\end{tabular}%
}
\caption{Ablation study with other VLMs replacing Opus 5.5 in our method.}
\label{tab:suppl_ablations}
\end{table}

%% file: sec_arxiv/tables/tab_vlm_variance.tex
\begin{table}[t]
\centering
\small
\setlength{\tabcolsep}{3pt}
\resizebox{\columnwidth}{!}{%
\begin{tabular}{l cccc cccc}
\toprule
 & \multicolumn{4}{c}{Prompt adherence $\uparrow$} & \multicolumn{4}{c}{Design principles $\uparrow$} \\
\cmidrule(lr){2-5} \cmidrule(lr){6-9}
Method & R1 & R2 & R3 & Avg & R1 & R2 & R3 & Avg \\
\midrule
Opus-Agent & 7.17 & 7.23 & 7.17 & 7.19 {\scriptsize$\pm$0.04} & 7.52 & 7.48 & 7.50 & 7.50 {\scriptsize$\pm$0.02} \\
\OURS{} (Ours) & \textbf{7.37} & \textbf{7.47} & \textbf{7.40} & \textbf{7.41} {\scriptsize$\pm$0.05} & \textbf{7.65} & \textbf{7.66} & \textbf{7.71} & \textbf{7.67} {\scriptsize$\pm$0.03} \\
\bottomrule
\end{tabular}%
}
\caption{Variance of the VLM evaluation on the 15-scene SAGE ablation set over 3 runs with different seeds. Each score is the mean of Qwen3-VL and Gemma 4 scores, averaged over all scenes. Avg is the mean over the three runs. The best result in each column is \textbf{bolded}.}
\label{tab:vlm_variance}
\end{table}

%% file: sec_arxiv/tables/tab_vlm_agreement.tex
\begin{table}[t]
\centering
\small
\begin{tabular}{lcc}
\toprule
Dataset & PA $\tau$ $\uparrow$ & DP $\tau$ $\uparrow$ \\
\midrule
SAGE & 0.78 & 0.90 \\
SAGE-Opus & 0.71 & 0.81 \\
SceneSmith & 0.88 & 0.90 \\
\bottomrule
\end{tabular}
\caption{Agreement between the two VLM judges (Qwen3-VL and Gemma~4): Kendall rank correlation $\tau$ between the rankings of the seven methods by their mean prompt adherence (PA) and design principles (DP) scores under each VLM.}
\label{tab:vlm_agreement}
\end{table}

%% file: sec_arxiv/fig_vlm_agreement.tex
\begin{figure*}[t]
\centering
\includegraphics[width=\textwidth]{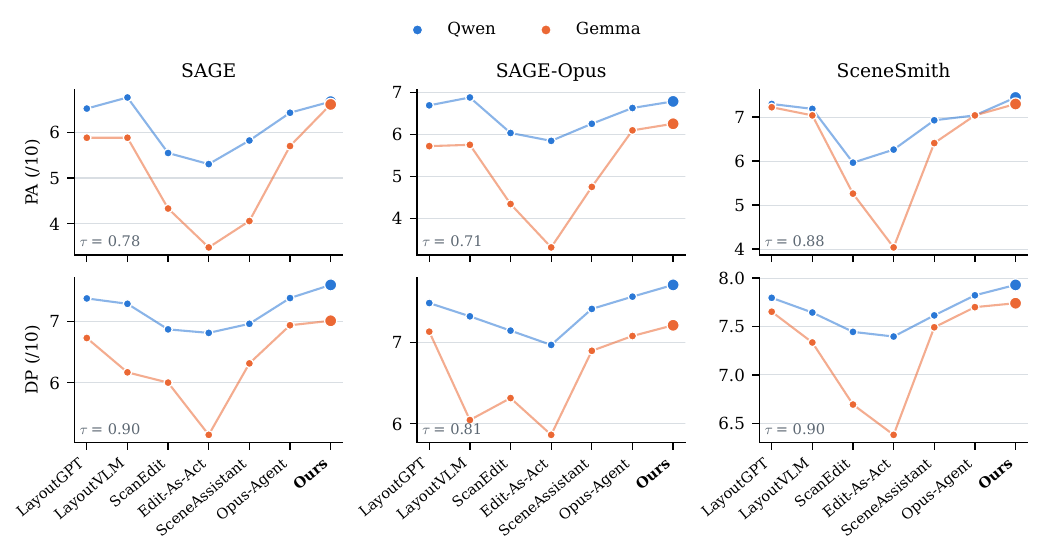}
\caption{Mean prompt adherence (PA, top) and design principles (DP, bottom) scores of each method under the two VLM judges, Qwen3-VL and Gemma~4, on SAGE, SAGE-Opus and SceneSmith. Both VLMs follow the same trend across methods; $\tau$ is the Kendall rank correlation between their method rankings.}
\label{fig:vlm_agreement}
\end{figure*}

%% file: sec_arxiv/fig_qualitative_suppl.tex
\begin{figure*}[!tp]
\centering
\setlength{\tabcolsep}{3pt}
\renewcommand{\arraystretch}{0.6}
\begin{tabular}{cccccc}
 & \parbox[b]{0.1881\textwidth}{\centering\scriptsize\itshape ``A co-working lounge with small desks for working as well as a corner to chat casually.''} & \parbox[b]{0.2203\textwidth}{\centering\scriptsize\itshape ``A lively after-work cocktail lounge where the bar is the social magnet and guests drift between standing chats near drinks and short perches nearby, rather than sitting down to a formal meal.''} & \parbox[b]{0.1540\textwidth}{\centering\scriptsize\itshape ``Turn the room into a small retro diner dance night by pushing the stools and other objects to the edges of the room.''} & \parbox[b]{0.1789\textwidth}{\centering\scriptsize\itshape ``A music rehearsal with one chair for the conductor and the rest comfortably spread out.''} & \parbox[b]{0.1487\textwidth}{\centering\scriptsize\itshape ``A cozy kitchen for two people to sit and eat while talking to a third person who is cooking.''} \\[2pt]
\rotatebox[origin=c]{90}{\parbox{0.13\textwidth}{\centering\scriptsize Input}} & \raisebox{-.5\height}{\includegraphics[width=0.1881\textwidth]{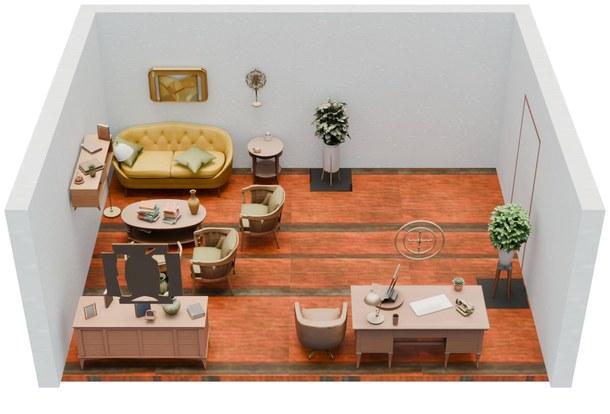}} & \raisebox{-.5\height}{\includegraphics[width=0.2203\textwidth]{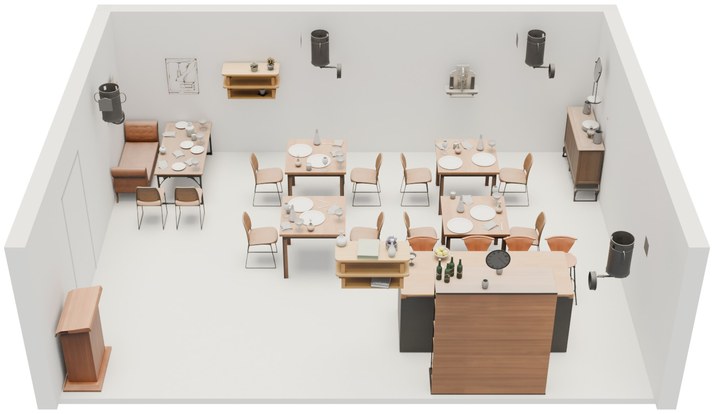}} & \raisebox{-.5\height}{\includegraphics[width=0.1540\textwidth]{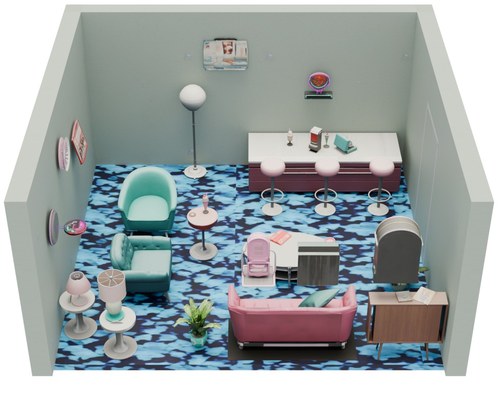}} & \raisebox{-.5\height}{\includegraphics[width=0.1789\textwidth]{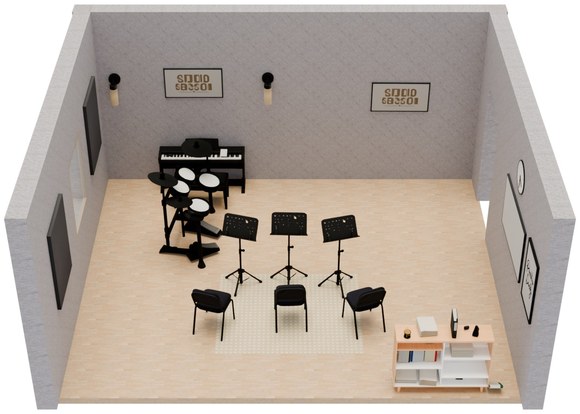}} & \raisebox{-.5\height}{\includegraphics[width=0.1487\textwidth]{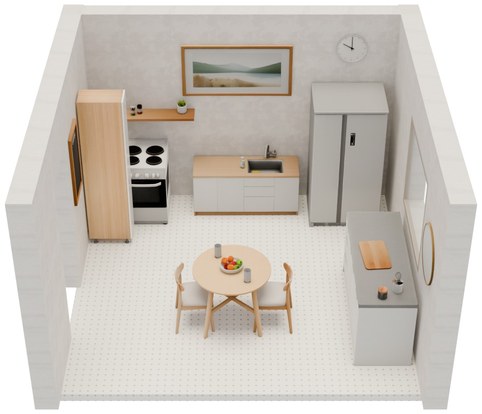}} \\
\rotatebox[origin=c]{90}{\parbox{0.13\textwidth}{\centering\scriptsize LayoutGPT}} & \raisebox{-.5\height}{\includegraphics[width=0.1881\textwidth]{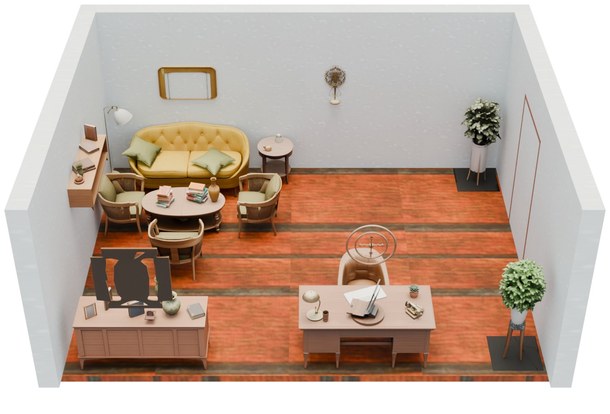}} & \raisebox{-.5\height}{\includegraphics[width=0.2203\textwidth]{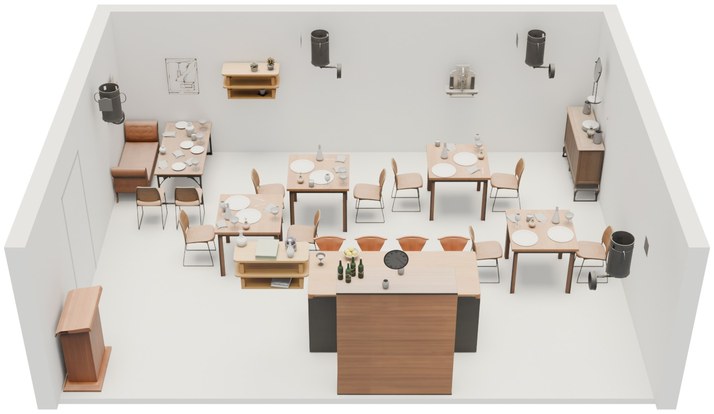}} & \raisebox{-.5\height}{\includegraphics[width=0.1540\textwidth]{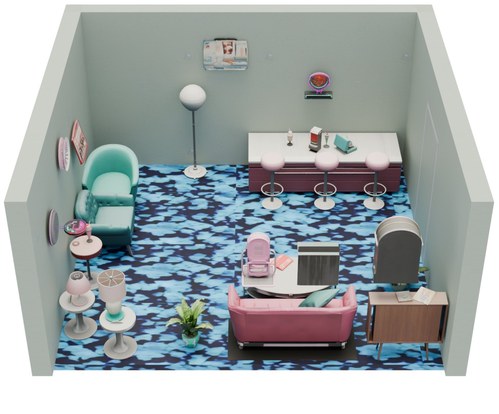}} & \raisebox{-.5\height}{\includegraphics[width=0.1789\textwidth]{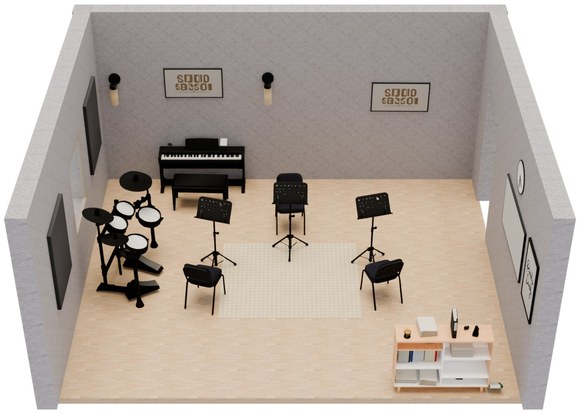}} & \raisebox{-.5\height}{\includegraphics[width=0.1487\textwidth]{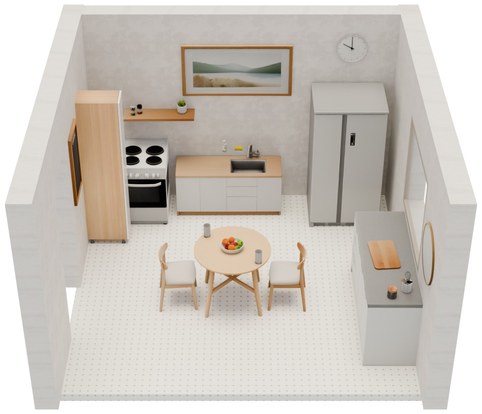}} \\
\rotatebox[origin=c]{90}{\parbox{0.13\textwidth}{\centering\scriptsize LayoutVLM}} & \raisebox{-.5\height}{\includegraphics[width=0.1881\textwidth]{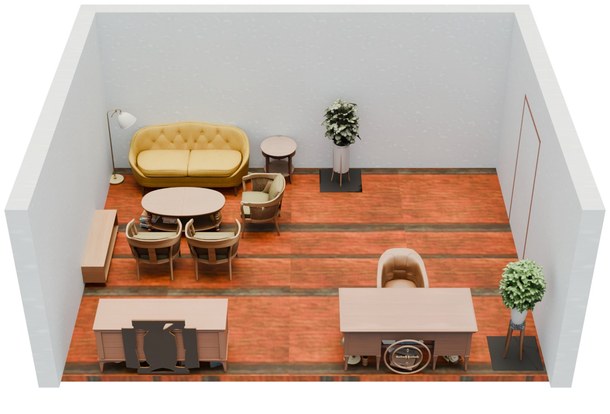}} & \raisebox{-.5\height}{\includegraphics[width=0.2203\textwidth]{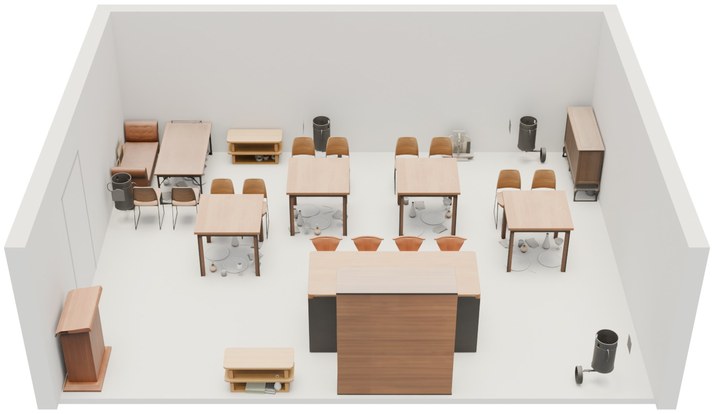}} & \raisebox{-.5\height}{\includegraphics[width=0.1540\textwidth]{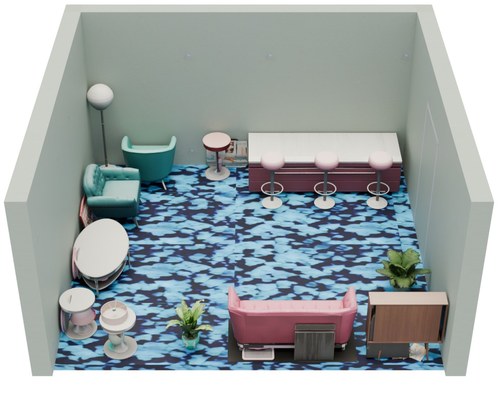}} & \raisebox{-.5\height}{\includegraphics[width=0.1789\textwidth]{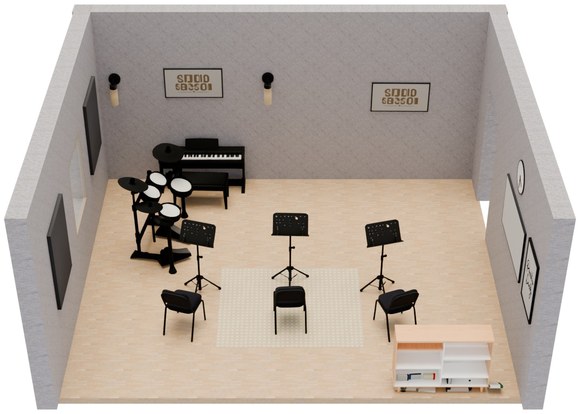}} & \raisebox{-.5\height}{\includegraphics[width=0.1487\textwidth]{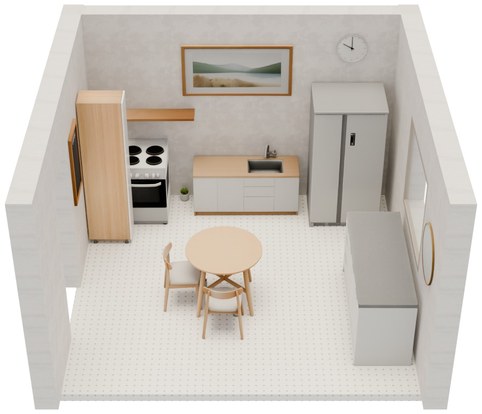}} \\
\rotatebox[origin=c]{90}{\parbox{0.13\textwidth}{\centering\scriptsize ScanEdit}} & \raisebox{-.5\height}{\includegraphics[width=0.1881\textwidth]{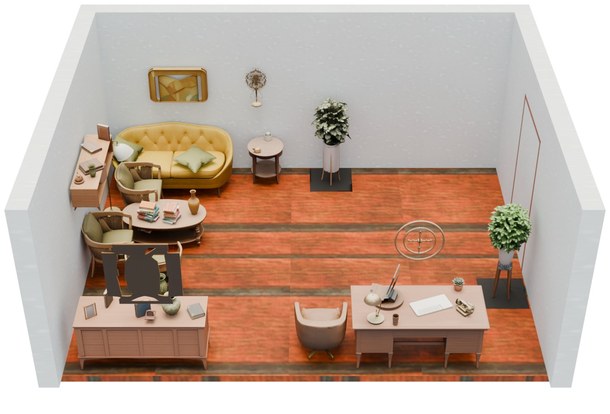}} & \raisebox{-.5\height}{\includegraphics[width=0.2203\textwidth]{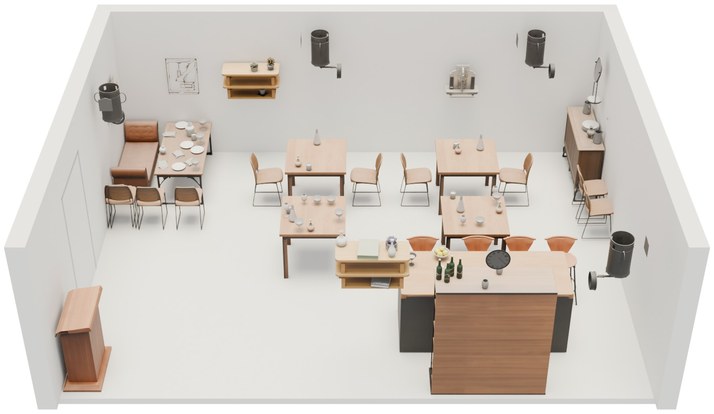}} & \raisebox{-.5\height}{\includegraphics[width=0.1540\textwidth]{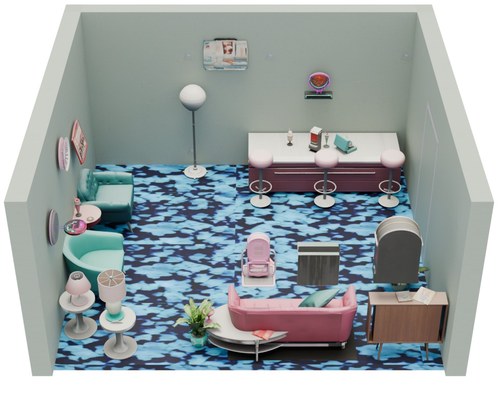}} & \raisebox{-.5\height}{\includegraphics[width=0.1789\textwidth]{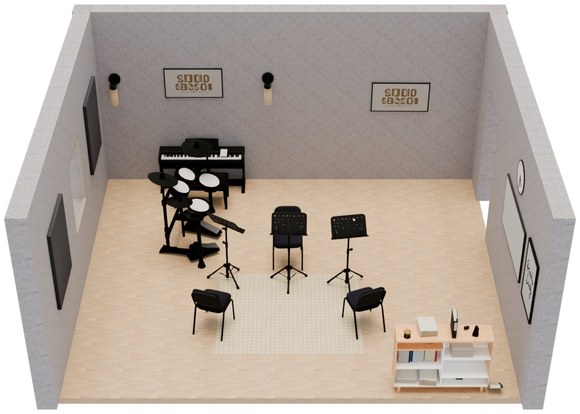}} & \raisebox{-.5\height}{\includegraphics[width=0.1487\textwidth]{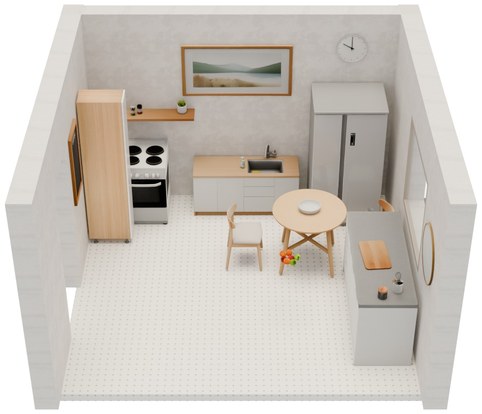}} \\
\rotatebox[origin=c]{90}{\parbox{0.13\textwidth}{\centering\scriptsize Edit-As-Act}} & \raisebox{-.5\height}{\includegraphics[width=0.1881\textwidth]{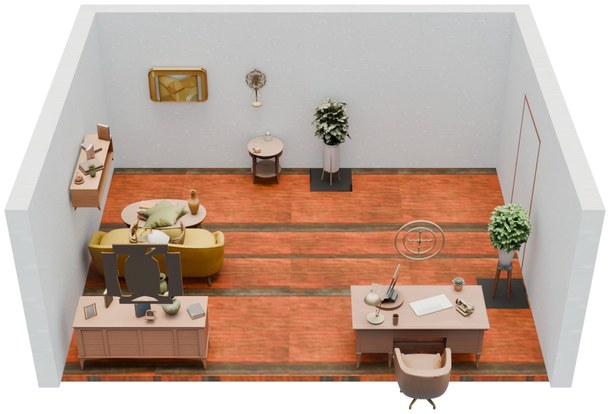}} & \raisebox{-.5\height}{\includegraphics[width=0.2203\textwidth]{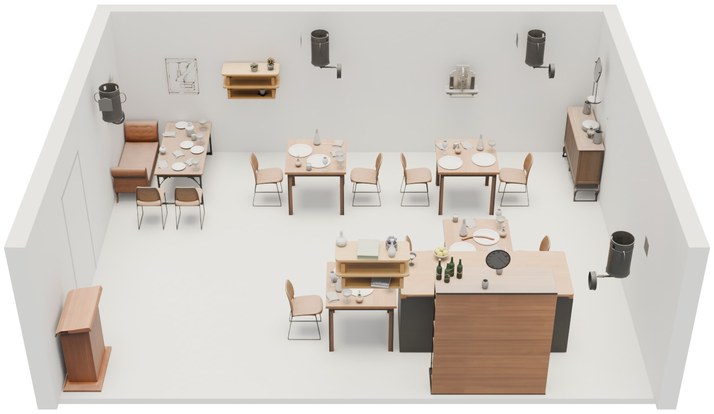}} & \raisebox{-.5\height}{\includegraphics[width=0.1540\textwidth]{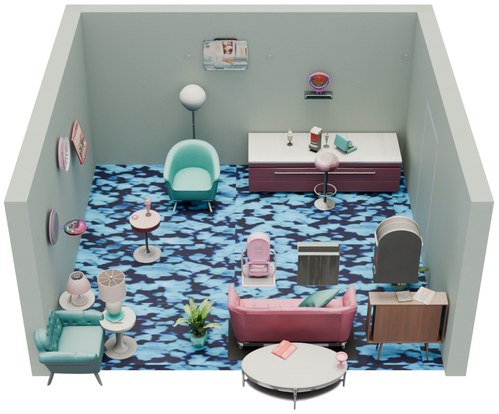}} & \raisebox{-.5\height}{\includegraphics[width=0.1789\textwidth]{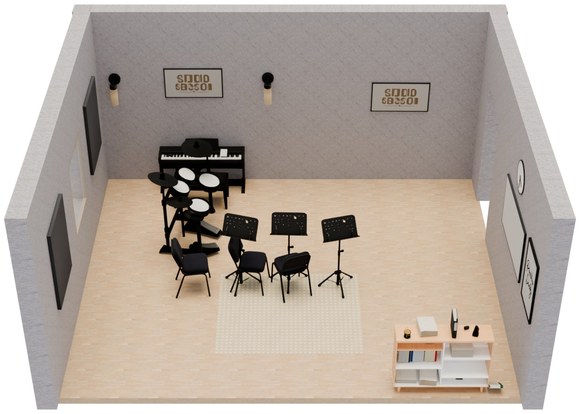}} & \raisebox{-.5\height}{\includegraphics[width=0.1487\textwidth]{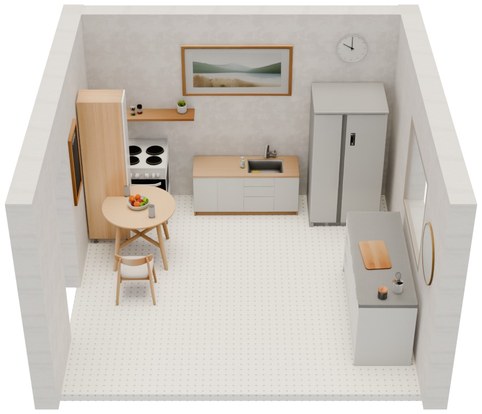}} \\
\rotatebox[origin=c]{90}{\parbox{0.13\textwidth}{\centering\scriptsize SceneAssistant}} & \raisebox{-.5\height}{\includegraphics[width=0.1881\textwidth]{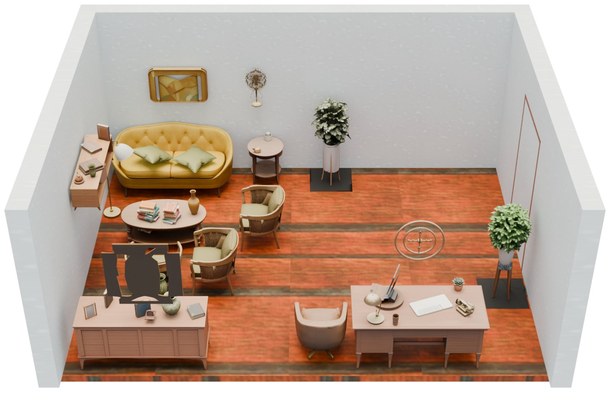}} & \raisebox{-.5\height}{\includegraphics[width=0.2203\textwidth]{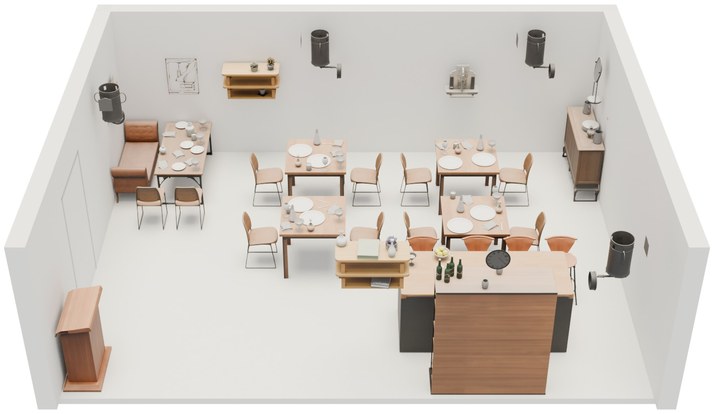}} & \raisebox{-.5\height}{\includegraphics[width=0.1540\textwidth]{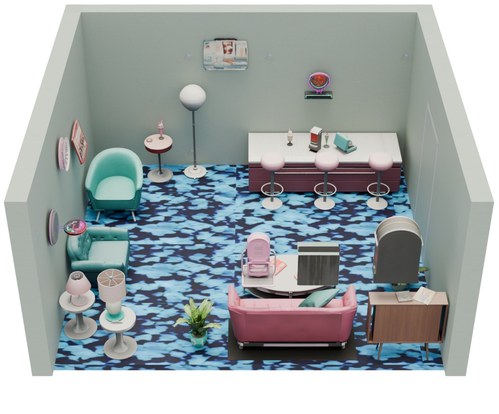}} & \raisebox{-.5\height}{\includegraphics[width=0.1789\textwidth]{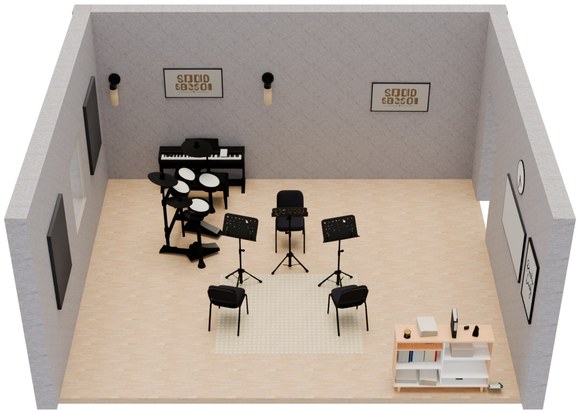}} & \raisebox{-.5\height}{\includegraphics[width=0.1487\textwidth]{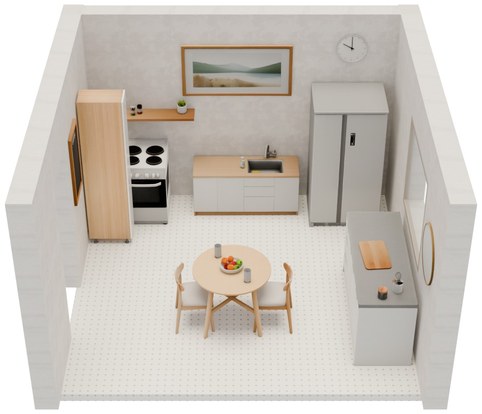}} \\
\rotatebox[origin=c]{90}{\parbox{0.13\textwidth}{\centering\scriptsize Opus-Agent}} & \raisebox{-.5\height}{\includegraphics[width=0.1881\textwidth]{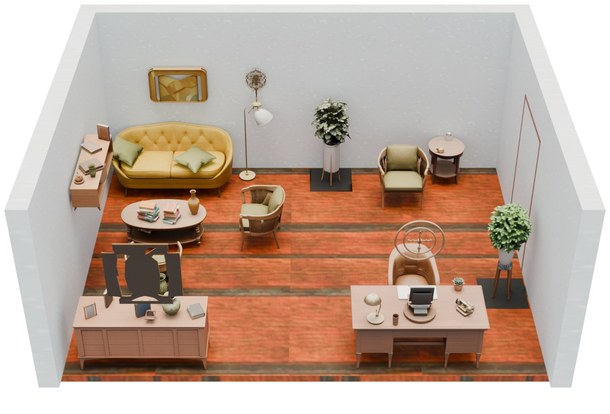}} & \raisebox{-.5\height}{\includegraphics[width=0.2203\textwidth]{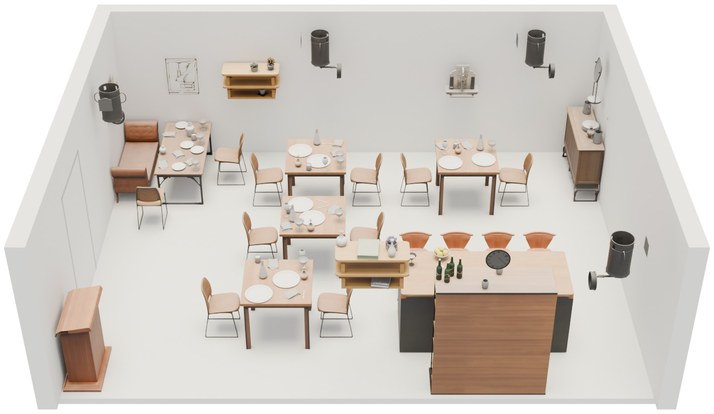}} & \raisebox{-.5\height}{\includegraphics[width=0.1540\textwidth]{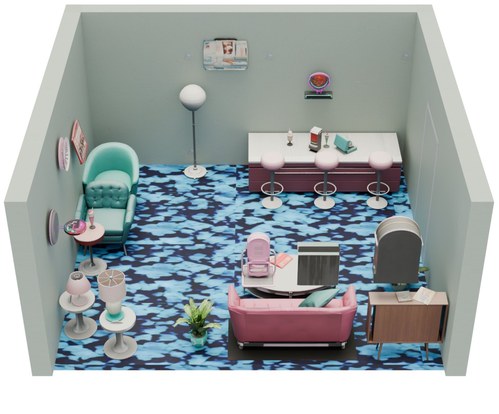}} & \raisebox{-.5\height}{\includegraphics[width=0.1789\textwidth]{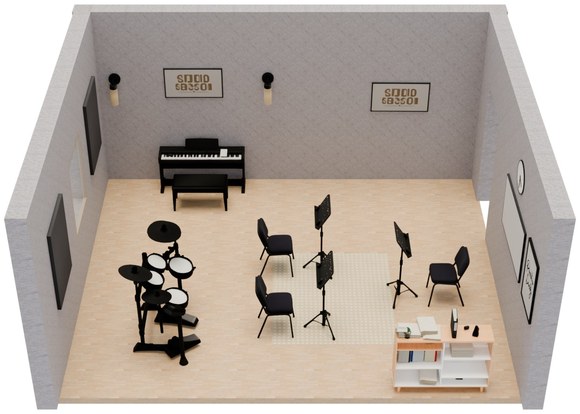}} & \raisebox{-.5\height}{\includegraphics[width=0.1487\textwidth]{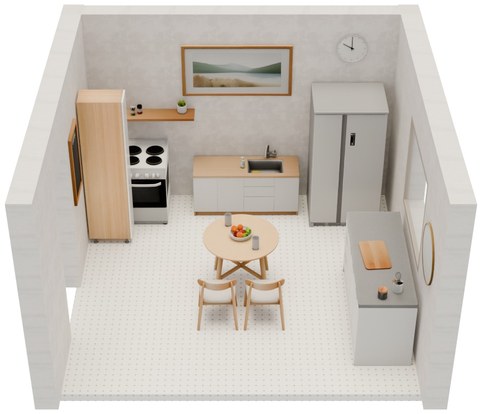}} \\
\rotatebox[origin=c]{90}{\parbox{0.13\textwidth}{\centering\scriptsize \OURS{}}} & \raisebox{-.5\height}{\includegraphics[width=0.1881\textwidth]{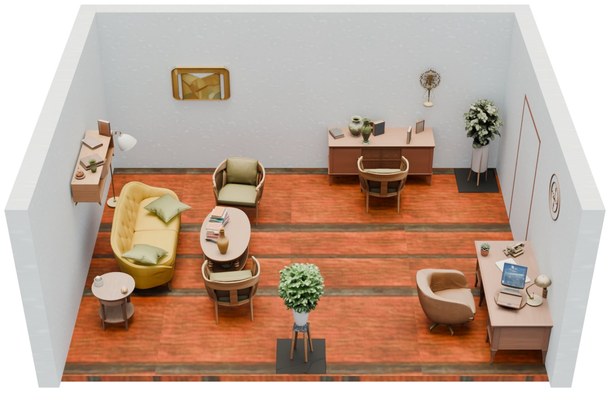}} & \raisebox{-.5\height}{\includegraphics[width=0.2203\textwidth]{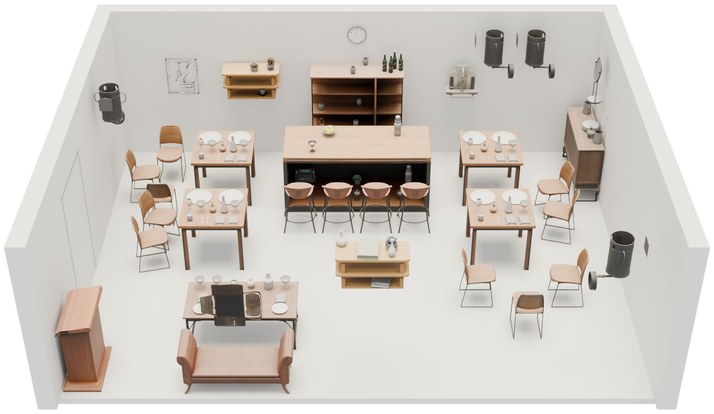}} & \raisebox{-.5\height}{\includegraphics[width=0.1540\textwidth]{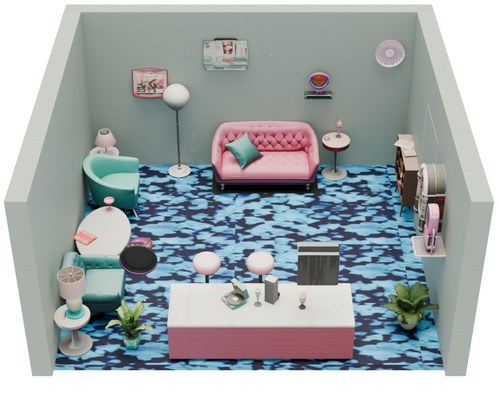}} & \raisebox{-.5\height}{\includegraphics[width=0.1789\textwidth]{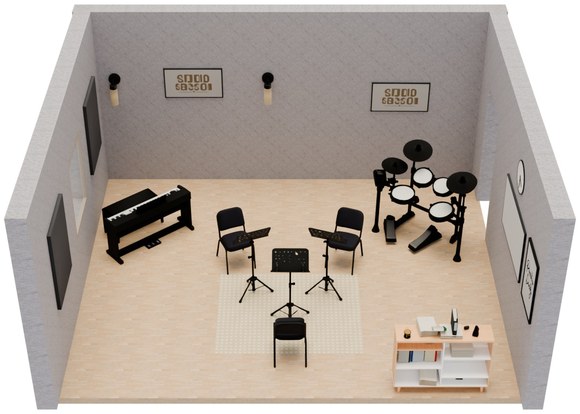}} & \raisebox{-.5\height}{\includegraphics[width=0.1487\textwidth]{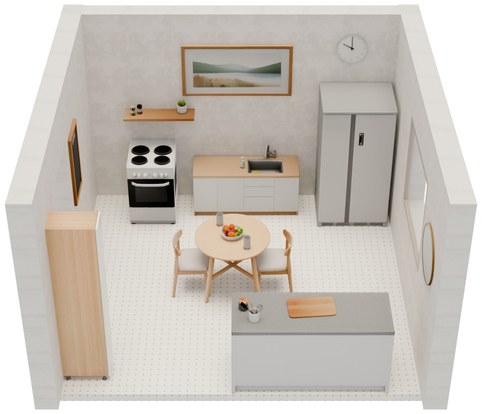}} \\
\end{tabular}
\caption{Additional qualitative comparison of \OURS{} against baselines on SAGE-Opus and SceneSmith. \OURS{} generates functional and aesthetic rearrangements.}
\label{fig:qual_suppl}
\end{figure*}

%% file: sec_arxiv/fig_diversity_suppl.tex
\begin{figure*}[!tp]
\centering
\setlength{\tabcolsep}{3pt}
\renewcommand{\arraystretch}{0.6}
\begin{tabular}{ccc}
 & \scriptsize Input & \begin{tabular}{@{}ccc@{}}\parbox{0.215\textwidth}{\centering\scriptsize Output 1}&\parbox{0.215\textwidth}{\centering\scriptsize Output 2}&\parbox{0.215\textwidth}{\centering\scriptsize Output 3}\end{tabular} \\[2pt]
\multicolumn{3}{c}{\parbox{0.860\textwidth}{\centering\scriptsize\itshape ``A social couch-gaming lounge where friends can play together, with everyone comfortably able to see the action and hear the sound while snacks and controllers pass around easily.''}} \\[2pt]
\begin{tabular}{@{}c@{}}\raisebox{-.5\height}{\rotatebox{90}{\makebox[0.1732\textwidth]{\scriptsize Opus-Agent}}}\\\raisebox{-.5\height}{\rotatebox{90}{\makebox[0.1732\textwidth]{\scriptsize \OURS{}}}}\end{tabular} & \begin{tabular}{@{}c@{}}\raisebox{-.5\height}{\includegraphics[width=0.215\textwidth]{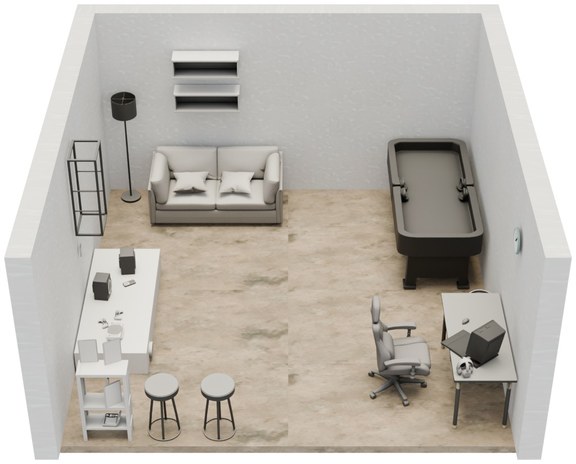}}\end{tabular} & \begin{tabular}{@{}ccc@{}}\raisebox{-.5\height}{\includegraphics[width=0.215\textwidth]{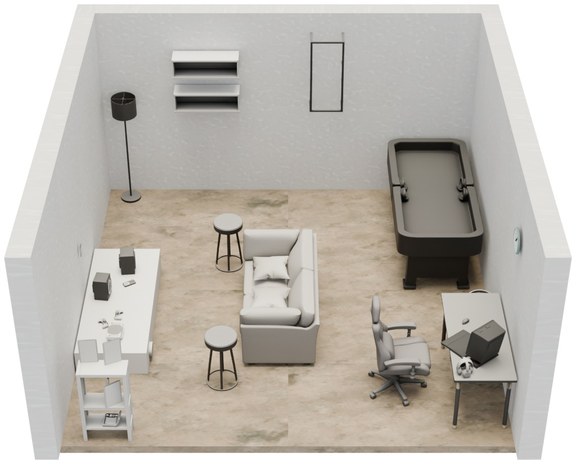}}&\raisebox{-.5\height}{\includegraphics[width=0.215\textwidth]{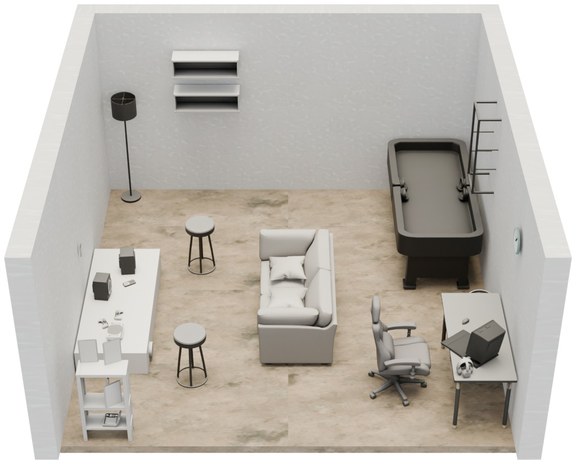}}&\raisebox{-.5\height}{\includegraphics[width=0.215\textwidth]{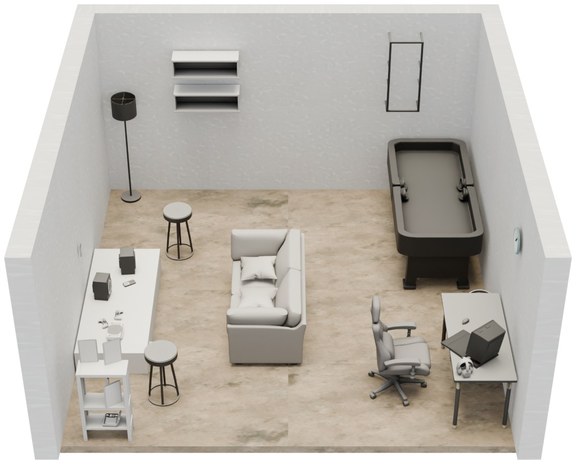}}\\\raisebox{-.5\height}{\includegraphics[width=0.215\textwidth]{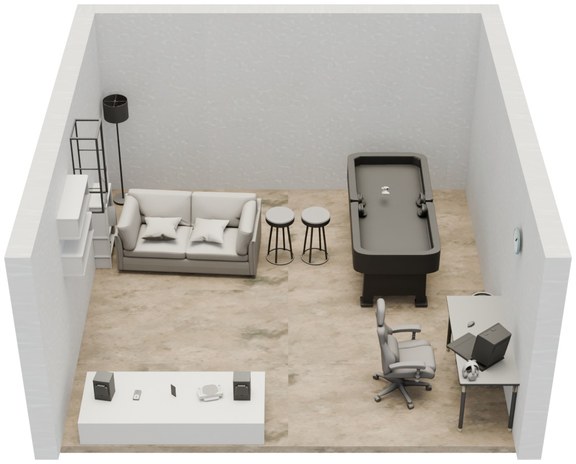}}&\raisebox{-.5\height}{\includegraphics[width=0.215\textwidth]{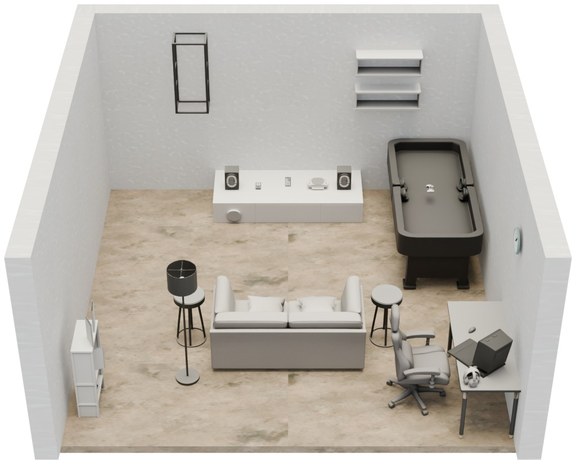}}&\raisebox{-.5\height}{\includegraphics[width=0.215\textwidth]{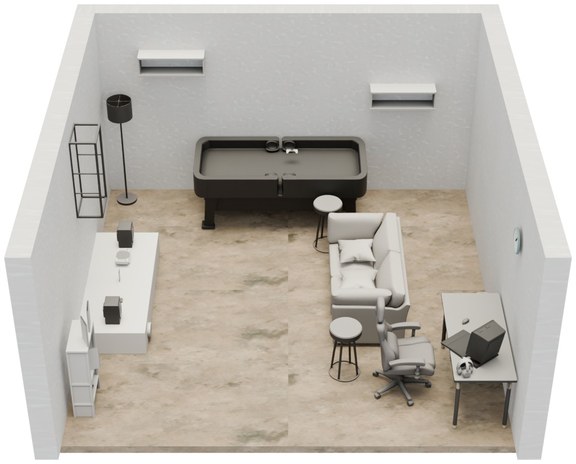}}\end{tabular} \\
\midrule
\multicolumn{3}{c}{\parbox{0.860\textwidth}{\centering\scriptsize\itshape ``A baby's room with many plants and soft lighting for calm sleep.''}} \\[2pt]
\begin{tabular}{@{}c@{}}\raisebox{-.5\height}{\rotatebox{90}{\makebox[0.1540\textwidth]{\scriptsize Opus-Agent}}}\\\raisebox{-.5\height}{\rotatebox{90}{\makebox[0.1540\textwidth]{\scriptsize \OURS{}}}}\end{tabular} & \begin{tabular}{@{}c@{}}\raisebox{-.5\height}{\includegraphics[width=0.215\textwidth]{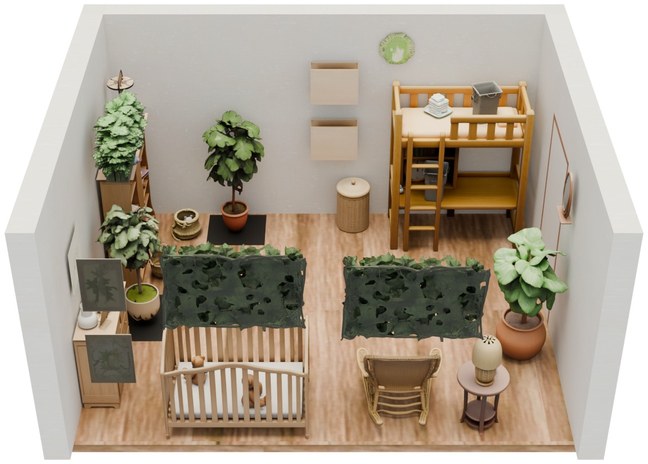}}\end{tabular} & \begin{tabular}{@{}ccc@{}}\raisebox{-.5\height}{\includegraphics[width=0.215\textwidth]{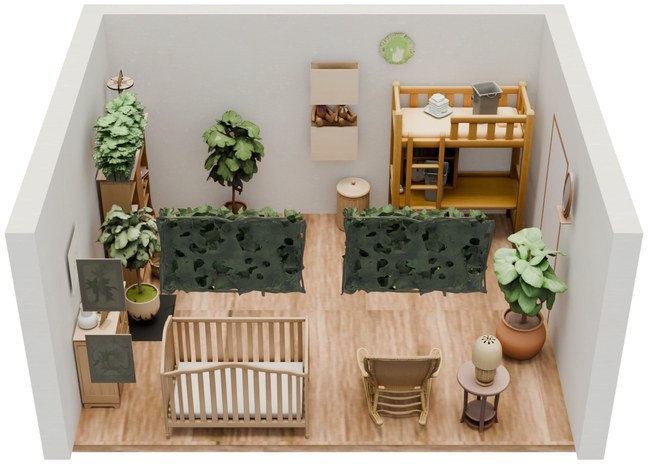}}&\raisebox{-.5\height}{\includegraphics[width=0.215\textwidth]{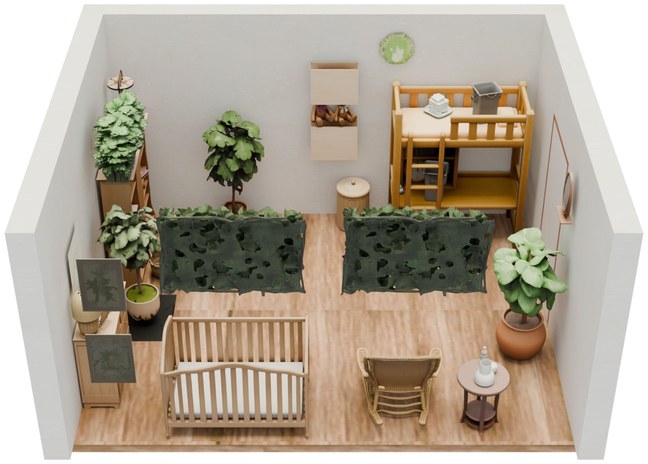}}&\raisebox{-.5\height}{\includegraphics[width=0.215\textwidth]{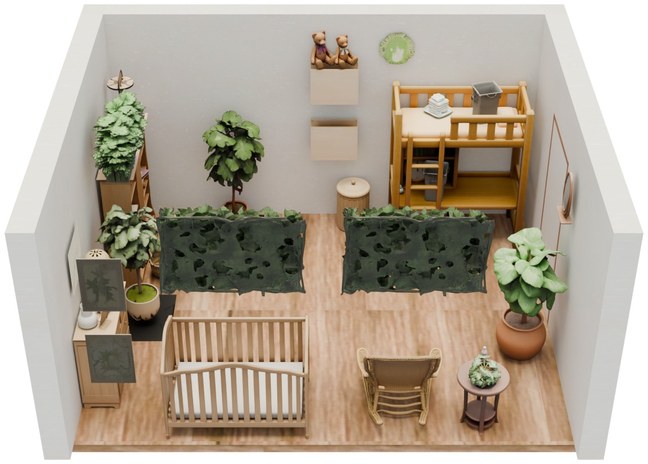}}\\\raisebox{-.5\height}{\includegraphics[width=0.215\textwidth]{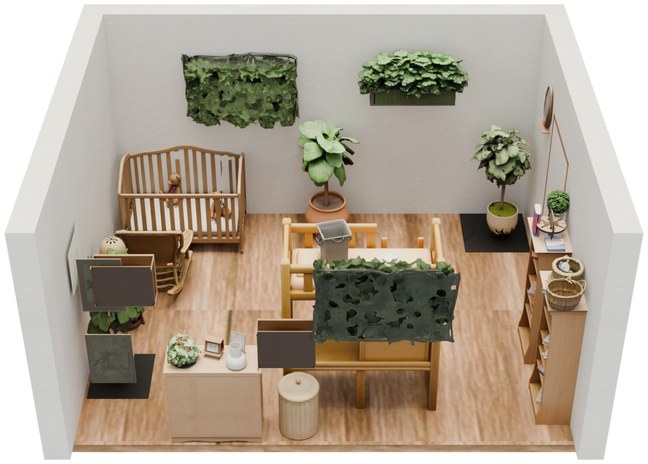}}&\raisebox{-.5\height}{\includegraphics[width=0.215\textwidth]{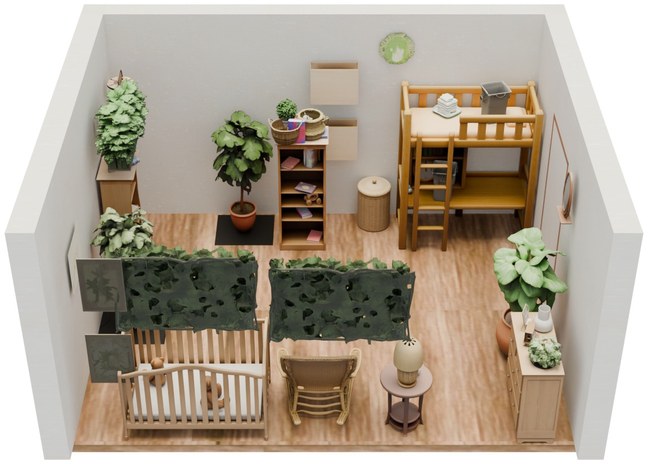}}&\raisebox{-.5\height}{\includegraphics[width=0.215\textwidth]{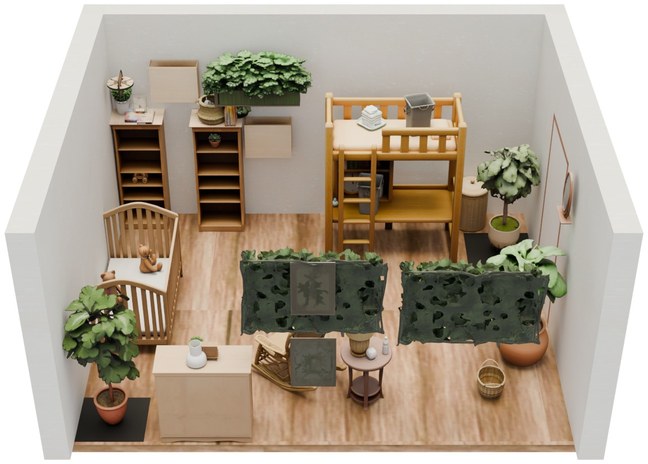}}\end{tabular} \\
\end{tabular}
\caption{Diversity of generated rearrangements. For each input scene and prompt, we compare three generated rearrangements from Opus-Agent and \OURS{}. \OURS{} generates more diverse rearrangements.}
\label{fig:div_suppl}
\end{figure*}